\documentclass[lettersize,journal]{IEEEtran}
\usepackage{amsmath,amsfonts}
\usepackage{algorithmic}
\usepackage{algorithm}
\usepackage{array}
\usepackage[caption=false,font=normalsize,labelfont=sf,textfont=sf]{subfig}
\usepackage{textcomp}
\usepackage{stfloats}
\usepackage{hyperref}
\usepackage{url}
\usepackage{verbatim}
\usepackage{graphicx}
\usepackage{cite}
\usepackage{multirow}
\usepackage{multicol}
\usepackage[table]{xcolor}
\usepackage{booktabs}
\usepackage{amssymb}
\newcommand{\cmark}{\checkmark}

\begin{document}

\title{From Cues to Horizons: Dynamic Risk Horizon Profiling \\ for Trajectory Prediction}


\author{
\IEEEauthorblockN{Xinyi Ning\textsuperscript{1}, Zilin Bian\textsuperscript{2,*},~\IEEEmembership{Member,~IEEE,}, Dachuan Zuo\textsuperscript{1}, Semiha Ergan\textsuperscript{1,*}, ~\IEEEmembership{Member,~IEEE,}, Kaan Ozbay\textsuperscript{1},~\IEEEmembership{Senior Member,~IEEE,}}
\thanks{\textsuperscript{1}Department of Civil and Urban Engineering, New York University. \textsuperscript{2} Department of Civil Engineering Technology and Environmental Management Safety, Rochester Institute of Technology.}
\thanks{* Corresponding Author: {\tt\small zilin.bian@rit.edu, semiha@nyu.edu}}
}


\markboth{Ning \MakeLowercase{\textit{et al.}}: From Cues to Horizons: Dynamic Risk Horizon Profiling for Trajectory Prediction}%
{Ning \MakeLowercase{\textit{et al.}}: From Cues to Horizons: Dynamic Risk Horizon Profiling for Trajectory Prediction}


\maketitle

\begin{abstract}
Accurate and reliable vehicle trajectory prediction is essential for safe autonomous driving. Recent studies have incorporated safety risk into trajectory prediction to quantify dangers posed by surrounding agents. However, most risk-aware approaches use past risk information as a secondary signal to help guide decisions, overlooking its future evolution and uncertainty. In this paper, we propose a risk horizon profiling (RHP) module that incorporates a continuous, learnable potential field model for risk-aware trajectory prediction. The RHP module calculates the spatial-temporal proximity of surrounding objects to profile risk distributions across future horizons, which supports better trajectory prediction by adaptively identifying what human drivers perceive as critical moments. We evaluate our method on two datasets from different driving settings, highD for highway corridors and SHRP2 for urban streets, which cover diverse risk scenarios including safe, near-crash, and crash events. Compared to the baseline methods, our framework achieves a 25.0\% reduction in 5s RMSE on the highD dataset and a 29.1\% reduction in 5s minFDE on SHRP2. These results indicate strong performance for both short and long horizon prediction and robust generalization across highway and urban scenarios. The proposed method enables more realistic AV path planning and strategic selection, thereby supporting safer autonomous driving and more advanced driver-assistance systems. The source code for this work is available at: \href{https://github.com/bilab-nyu/RHP}{link}
\end{abstract}


\section{Introduction}
The safety of autonomous vehicles (AVs) is a critical concern that directly impacts their public acceptance and large-scale deployment \cite{bharilya2024machine}. A key component is accurate vehicle trajectory prediction, which forecasts the future positions of vehicles and guides the AV’s motion planner to identify low-risk paths \cite{huang2025post}. It is usually framed as a time-series forecasting problem, employing data-driven methods to capture trajectory patterns and social interactions \cite{li2019grip++, zhang2022ai, gao2025ise}. More recently, studies have shifted toward a human-centered approach, incorporating human cognition into predictive models for more accurate and naturalistic predictions \cite{li2024context, liao2024bat, gao2023dual}.

\begin{figure}[!t]
\centering
\includegraphics[width=\columnwidth]{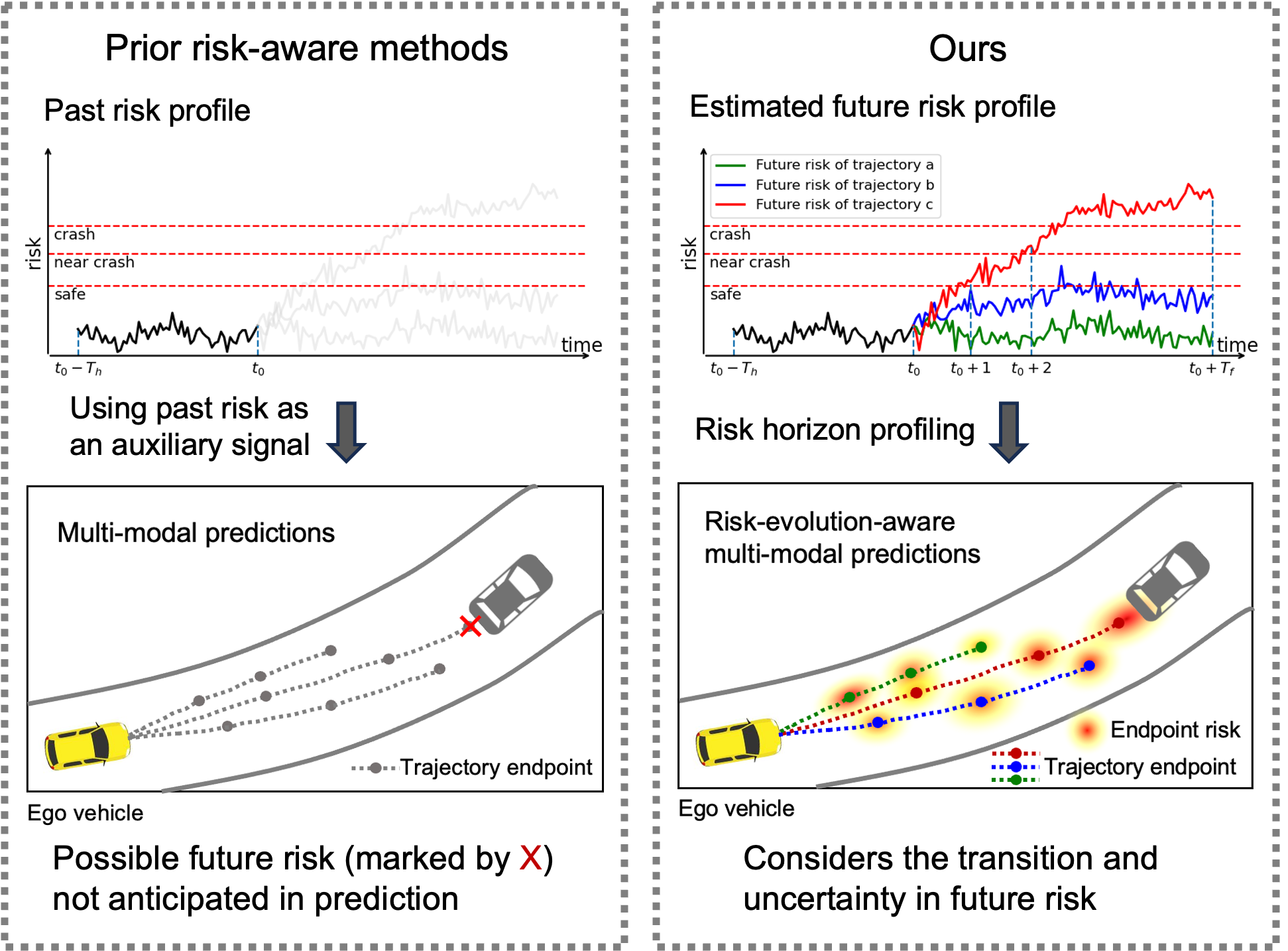}
\caption{Conceptual comparison of risk-aware trajectory prediction. (Left) Prior methods compute a past risk profile over the history window $[t_0-T_h,t_0]$ and append it as an auxiliary signal, which can overlook potential collisions among candidate trajectories. (Right) Our framework estimates multi-horizon future risk profiles over $[t_0,t_0+T_f]$, enabling risk-evolution-aware multi-modal predictions that consider the spatial-temporal evolution and uncertainty of risk. }
\label{vision}
\end{figure}

Among various aspects of human cognition, safety risk has become an important consideration in trajectory prediction, offering an interpretable metric of potential dangers from uncertain or aggressive surrounding agents, yet existing risk-aware approaches still face several limitations \cite{zhu2022interaction, zhu2025incorporating}. Most methods compute risk variations only from past observations and incorporate them as auxiliary signals \cite{wang2025risk}, without explicitly modeling the uncertainty and transitions of risk in the future spatial-temporal domain, as illustrated in Fig. \ref{vision}. In practice, the future risk of a vehicle can vary across candidate trajectories and over time. One trajectory may remain safe while another leads to a collision, and the same pair of vehicles may shift from safe interaction to near-crash and then diverge again within seconds. Such spatial-temporal fluctuations of risk should be accounted for during trajectory prediction. Relying solely on historical risk creates a risk–horizon mismatch by overlooking how future risk is distributed across trajectories and time horizons, making it difficult for the model to identify the most critical future time points.

Furthermore, prior risk-aware methods rarely consider the scenario dependence in drivers’ attention to risk. Under safe conditions with relatively stable speed variations, drivers tend to rely on reflexive control and prioritize imminent, short-horizon risk cues. In contrast, in high-risk situations, their attention may shift toward longer-horizon risk patterns several seconds ahead, since short-term reactive responses may be insufficient to ensure safety and future states must be proactively anticipated. These preferences may be implicitly reflected in trajectory data and could be leveraged to guide context-dependent trajectory generation under crash, near-crash, and safe scenarios.

To address these issues, we propose a risk-evolution-aware trajectory prediction framework based on dynamic risk horizon profiling. For risk quantification, we adopt a continuous and learnable risk potential field model that jointly captures spatial proximity to surrounding objects and the temporal proximity of potential collisions. For risk horizon profiling, we introduce an auxiliary coarse endpoint prediction task for all vehicles and use the predicted endpoints to estimate the risk distribution at each prediction horizon. We further design a horizon-importance attention mechanism to assign adaptive weights to each future horizon, emphasizing the most critical future moments for crash, near-crash, and safe scenarios. The resulting multi-horizon risk profile captures not only the magnitude of risk, but also how risk evolves across candidate trajectories over time. Finally, we integrate the risk profile with vehicle kinematic and interaction features for final trajectory prediction. 

The main contributions of this paper are summarized as follows.

\begin{itemize}
    \item We propose a dynamic risk horizon profiling module for risk-evolution-aware trajectory prediction, capturing the uncertainty and evolution of future risk across candidate trajectories and prediction horizons.
    \item We adopt a learnable risk potential field to quantify the spatial-temporal proximity of surrounding vehicles, and design a horizon-importance attention mechanism to adaptively highlight critical future moments under different risk scenarios.
    \item Our model is validated on the highway dataset highD and the urban dataset SHRP2, achieving a 25.0\% reduction in RMSE on highD and a 29.1\% reduction in minFDE on SHRP2 over a 5-second prediction horizon compared with existing state-of-the-art prediction models, demonstrating its robustness and generalizability.
\end{itemize}

\section{Related Work}
\subsection{Trajectory Prediction}

The goal of trajectory prediction is to predict the future trajectories of the traffic agents from a brief history of their past movements. Many approaches treat this as a sequence modeling problem and use neural networks to learn temporal dynamics and spatial interactions from trajectory data. Commonly adopted ones include long short-term memory (LSTM) networks \cite{kim2017probabilistic, lin2021vehicle,weiss2026barte}, graph neural networks (GNNs) \cite{liang2020learning, li2020evolvegraph,chib2024ms}, transformers \cite{shi2022motion, cao2025fif}, generative adversarial networks (GANs) \cite{gupta2018social, zhao2020novel}, and diffusion models \cite{westny2024diffusion, yang2024intp}. 

Building on these models, many studies incorporate high-level driving goals. They either predict discrete intention labels \cite{chen2022vehicle} or generate a dense set of candidate destinations to represent all possible future endpoints of the ego vehicle to guide multi-modal trajectory prediction \cite{zhao2021tnt,chen2023goal,gu2021densetnt}. While such intention-based or goal-based methods provide a structured way to integrate intent into trajectory prediction, they often struggle to respond to risky scenarios such as collisions or cut-in maneuvers\cite{zhu2026scenefactory}, and discrete intention categories or candidate goals limit their ability to capture the continuous and dynamic nature of real-world driving behavior. 

\subsection{Risk-aware trajectory prediction}

To improve model’s ability to handle safety-critical scenarios, recent studies have shifted their focus toward risk-aware trajectory prediction. These methods assess potential threats from surrounding objects and incorporate them into predictive models to improve performance, especially under high-risk interactions. For instance, risk can be added as additional features in the input representations to capture hazards associated with nearby vehicles, lanes, and other road agents, enabling models to reason about safety through interactions \cite{liu2022interactive, wang2025risk, liao2025sa}. Risk can also be modeled in graphs via risk-informed relational edges \cite{fang2023heterogeneous, feng2025risk}. In addition, some studies treat risk as an optimization signal, integrating collision likelihood and severity into the loss to encourage safer predictions \cite{wei2024intention, dang2023ttc, liao2025sa} or improved accuracy in high-risk scenarios \cite{ning2025strap, thuremella2024risk}. Despite these advances, current risk-aware methods often treat risk as an auxiliary signal appended to the basic past-trajectory information, without modeling its future evolution or uncertainty, which limits their robustness and generalization across driving scenarios from safe to near-crash or crash.

\section{Methodology}

\begin{figure*}[!t]
\centering
\includegraphics[width=\textwidth]{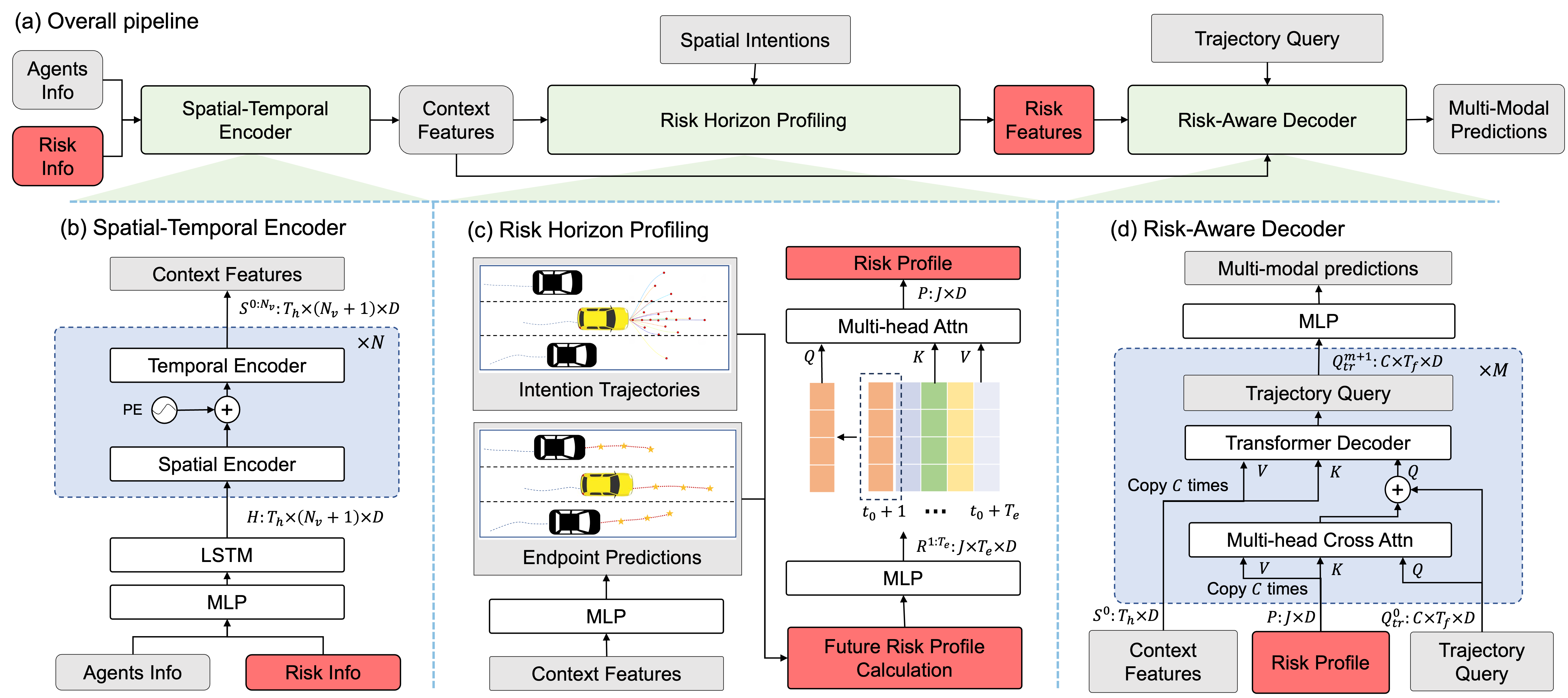}%
\caption{The architecture of the RHP framework. Panel a illustrates the pipeline of the framework, including a spatial-temporal encoder (panel b), a risk horizon profiling module (panel c), and a risk-aware decoder (panel d). The risk-related modules and variables are marked in \textcolor{red}{red} and the rest of the inputs and outputs are marked in \textcolor{gray}{gray}. }
\label{fig:architecture}
\end{figure*}

\textbf{Problem Formulation.} The inputs of trajectory prediction are past states of the ego vehicle and its neighbor vehicles from time step \(-T_h\) to 0. The states \(X\) include the x-y coordinates, velocity, acceleration and a perceived risk score computed by our risk potential field model to quantify interaction risk at each time step, as well as vehicle type and lane ID. The output is multi-modal, where the model produces \(C\) candidate future trajectories of the ego vehicle, including the velocities and x-y coordinates from time step \(1\) to \(T_f\). The velocity for trajectory mode \(c\) is given by \((\hat{v}_{x,t}^c, \hat{v}_{y,t}^c)\). For the 2D position of mode \(c\) at time \(t\), we use a bivariate Gaussian distribution parameterized by $Y_t^c=(\hat{\mu}_{x,t}^c,\hat{\mu}_{y,t}^c,\hat{\sigma}_{x,t}^c,\hat{\sigma}_{y,t}^c,\hat{\rho}_t^c)$. Each mode is also assigned a probability \(\hat{p}^c\). Consequently, the spatial occurrence probability \(P(u_t)\) of the ego vehicle at time \(t\) is computed as:
\begin{equation}
\label{probability}
P(u_t)=\sum_{c=1}^C\hat{p}^c\cdot {\mathcal{N}}\left(x_{t}-\hat{\mu}^c_{x,t},\hat{\sigma}^c_{x,t};y_{t}-\hat{\mu}^c_{y,t},\hat{\sigma}^c_{y,t};\hat{\rho}^c_t\right)
\end{equation}
where \(u_t=(x_t,y_t)\) is the x-y coordinate of the ego vehicle, \(C\) is the number of predicted trajectory modes and \(\sum_{c=1}^C\hat{p}^c=1\). 

\textbf{Overall Framework.} The overall methodological pipeline is illustrated in Fig. \ref{fig:architecture}. The spatial-temporal encoder takes the agent information as input, including the states of the ego vehicle and its neighboring agents, along with historical risk values. The risk horizon profiling module leverages the encoded context features to generate a comprehensive future risk profile. Finally, the risk-aware decoder combines the risk profile and context features to predict multi-modal trajectories.

\subsection{Risk Potential field}

We adopt the continuous risk potential field model proposed by Zuo et al. \cite{zuo2025composite} for risk assessment. It comprises a subjective risk field that measures the spatial proximity of surrounding objects and an objective risk field that captures the temporal proximity to a potential collision.

\textbf{Definition 1} (Subjective risk)\textbf{.} The subjective risk field models the ego vehicle’s perceived risk based on its spatial proximity to surrounding vehicles. It assumes that the ego vehicle maintains a subjective safe distance from its surroundings and assigns higher risk values as the distance to nearby vehicles decreases. Let \(s^i=(x^i,y^i,v_x^i,v_y^i)\) denote the longitudinal and lateral position and velocity of vehicle \(i\). The subjective risk perceived by the ego vehicle \(i\) due to its spacing with another vehicle \(j\) can be formulated as:
\begin{equation}
r^{ij}_s=\exp\left(-\left|\frac{x^i-x^j}{\gamma_x}\right|^{\beta_x}-\left|\frac{y^i-y^j}{\gamma_y}\right|^{\beta_y}\right)
\label{eq:s-field}
\end{equation}
where $\gamma_x > 0$ and $\gamma_y > 0$ are the scaling factors, and $\beta_x \geq 2$ and $\beta_y \geq 2$ are the corresponding shape parameters. All four are learnable risk parameters and are optimized with the rest of the network via backpropagation during training.

\textbf{Definition 2} (Objective risk)\textbf{.} The objective risk field captures the temporal dimension of collision risk by estimating the likelihood of a collision during the relative motion of two vehicles. It jointly considers how close the vehicles are expected to become and how soon the closest encounter will occur. For example, in a cut-in scenario, the subjective risk field may remain moderate because the current distance is not yet small, whereas the objective risk field can be high due to the imminent collision tendency. The objective risk between vehicle \(i\) and vehicle \(j\) is calculated as:
\begin{equation}
t_m^{ij}=-\frac{(x^i-x^j)(v_x^i-v_x^j)+(y^i-y^j)(v_y^i-v_y^j)}{(v_x^i-v_x^j)^2+(v_y^i-v_y^j)^2}
\label{eq:tm}
\end{equation}
\begin{equation}
d_m^{ij}=\sqrt{\frac{\left[(y^i-y^j)(v_x^i-v_x^j)-(x^i-x^j)(v_y^i-v_y^j)\right]^2}{(v_x^i-v_x^j)^2+(v_y^i-v_y^j)^2}}
\label{eq:tm}
\end{equation}
\begin{equation}
r^{ij}_o=\exp\left[-\left(\frac{{d}_{m}^{ij}}{d^*}\right)^{\beta_p}\right]exp\left[-\left(\frac{{t}_{m}^{ij}}{t^*}\right)^{\beta_t}\right]
\label{eq:o-field}
\end{equation}
where \(t_m^{ij}\) is the time step at which the distance between \(i\) and \(j\) stops decreasing, and \(d^{ij}_m\) is their future minimum distance. Here, $d^*$ is a distance threshold that determines whether a collision occurs, $t^*$ is a temporal scaling factor, and $\beta_p$ and $\beta_t$ are the corresponding shape parameters. The parameters \((d^*, t^*, \beta_p, \beta_t)\) are all learnable parameters and are optimized during training.
 
The risk between two vehicles is represented as a vector pointing from the ego vehicle toward the neighboring vehicle that induces the risk. This vector is then projected onto the longitudinal and lateral axes of the ego vehicle's coordinate frame. By incorporating these directional risks, the model can distinguish whether the threat comes from the front, rear, left, or right, aligning more closely with real-world traffic dynamics.

\subsection{Spatial-Temporal Encoder}
The spatial-temporal encoder captures both the spatial relationships among interacting vehicles within a scene and the temporal interaction patterns of each vehicle across historical timestamps. The architecture is illustrated in Fig. \ref{fig:architecture}(b). The encoder first applies an MLP and an LSTM to convert the input agent states and associated risk signals into latent embedding \(H\in\mathbb{R}^{T_h\times (N_v+1)\times D}\), where \(T_h\) is the history time frames, \(N_v\) is the number of neighboring vehicles and \(D\) is the feature dimension.

For spatial encoding, we feed the feature vectors of all vehicles at each timestep into a multi-head self attention layer followed by a gated linear unit (GLU). A residual connection with layer normalization is applied to combine the spatial-interaction features with the original motion dynamics. Then, we perform temporal encoding by aggregating, for each vehicle, its features across all history steps to capture the temporal moving intentions of each vehicle. The temporal encoder adopts the same attention-GLU structure, with added positional encoding. By stacking the spatial and temporal encoder layers \(N\) times, the encoder learns deeper hierarchical context representations, denoted as \(S^{0:N_v}\in\mathbb{R}^{T_h\times (N_v+1)\times D}\).

\subsection{Risk Horizon Profiling}
The risk horizon profiling module estimates the risk evolution across all future horizons and generates a comprehensive risk profile that highlights driver-attended critical moments, as illustrated in Fig. \ref{fig:architecture}(c). 

To profile future risk evolution, we require estimates of future states of all vehicles over the prediction horizons. We therefore introduce an auxiliary coarse endpoint prediction task to predict the positions and velocities of each vehicle at the end of every horizon. Unlike the final trajectory prediction module, which outputs a full sequence at all time steps, this auxiliary task produces only horizon-end states. We take the encoder context features \(S^{0:N_v}\) and apply a lightweight MLP to obtain coarse future endpoints for each vehicle \(i\), denoted by \(\hat{G}^i=\left\{\hat{g}^i_1, \hat{g}^i_2, ..., \hat{g}^i_{T_e}\right\}\), where \(i\in[0, N_v]\), \(N_v)\) is the number of vehicles, and \(T_e\) is the number of prediction horizons. At each horizon endpoint \(t\), the state of vehicle \(i\) is \(\hat{g}^i_t=(\hat{x}_{t}^i,\hat{y}_{t}^i,\hat{v}_{x,t}^i,\hat{v}_{y,t}^i)\), where \(\hat{x}_{t}^i,\hat{y}_{t}^i\) are the predicted x-y coordinates, and \(\hat{v}_{x,t}^i,\hat{v}_{y,t}^i\) are the corresponding longitudinal and lateral velocities.

To account for the uncertainty of the ego vehicle’s intended destination, we apply k-means clustering to the ground-truth trajectories and extract \(J\) representative intention trajectories for the ego vehicle. Each intention trajectory \(j\) serves as an implicit motion mode and is described by \(I_{t}^j=(x_{t}^j,y_{t}^j,v_{x,t}^j,v_{y,t}^j)\), where \(j\in\left\{1,...,J\right\}\). These intention trajectories are clustered from the training dataset and shared across all samples during training and inference. At each horizon endpoint \(t\), risk is computed using the coarse endpoint predictions and these pre-clustered trajectories, with the risk \(R_t^j\) along the \(j\)-th intention trajectory defined in \eqref{eq:endpointrisk}.
\begin{equation}
R_t^j=\sum_{i=1}^{N_v}\mathcal{F}_{risk}(I^j_t,\hat{G}^i_t)
\label{eq:endpointrisk}
\end{equation}
where \(\mathcal{F}_{risk}\) can be the subjective or objective risk function presented in Section 3.1.

After obtaining the multi-horizon risk distributions, we introduce a horizon-importance attention mechanism to identify the future moments that are most relevant to trajectory prediction. Instead of directly feeding all risk estimates into the decoder or imposing handcrafted rules to determine which horizon is most critical, this module learns from trajectory data and adaptively aggregates risk information across horizons. For each intention mode, we first concatenate the intention trajectory with its corresponding risk values and pass them through an MLP to obtain risk embeddings \(R\in\mathbb{R}^{J\times T_e\times D}\), where \(D\) denotes the feature dimension. We then use the 1-second endpoint risk embedding \(R_1^{j}\) of each intention mode \(I^j\) as the attention query and compute cross-attention over the full horizon risk embeddings \(R_{1:T_e}^{j}\). The 1-second risk embedding is selected as the query because it represents a high-confidence and immediately controllable risk state, whereas longer-horizon risks are generally more uncertain and noisy. The resulting attention weights quantify the relative contribution of each future horizon to the aggregated risk feature, conditioned on the 1-second risk. The attention formulation is given in \eqref{eq:rhp}. 
\begin{equation}
P^{j}=\mathrm{CrAttn}(q=R_1^{j},k=R_{1:T_e}^{j},v=R_{1:T_e}^{j})
\label{eq:rhp}
\end{equation}
where \(P^{j}\in\mathbb{R}^{J\times D}\) is the aggregated risk feature for the \(j\)-th intention trajectory, and \(\mathrm{CrAttn}(·)\) represents the multi-head cross attention layer.


\subsection{Risk-Aware Decoder}
 The risk-aware decoder takes the context features and risk features as inputs, and outputs multi-modal trajectory predictions together with the probability of each mode, as illustrated in Fig. \ref{fig:architecture}(d). We adopt the method from autobot \cite{girgis2021latent} and introduce \(c\) sets of learnable seed parameters, referred to as trajectory query \(Q\in\mathbb{R}^{c\times T_f\times D}\), where \(T_f\) is the number of prediction time steps. By inferring \(c\) modes in parallel, the trajectory query enables efficient single-pass inference over the entire future scene. Specifically, the trajectory query maps risk and context features into distinct future trajectory candidates, and is iteratively updated through an $M$-layer transformer architecture. The initial query \(Q^0\) is initialized with Xavier Uniform Initialization. A multi-head cross attention module is first employed to embed the risk features into the trajectory query as follows:
\begin{equation}
{Q}^{m'}=\mathrm{CrAttn}(q=Q^m,k=\tilde{P},v=\tilde{P})
  \end{equation}
where \(\tilde{P}\in\mathbb{R}^{c\times J\times D}\) are the risk features from the risk profile decoder copied \(c\) times, \(Q^m\) is the trajectory query of the \(m\)-th layer, and \({Q}^{m'}\in\mathbb{R}^{c\times T_f\times D}\) is the output of the multi-head attention layer.

We then use a transformer decoder to further propagate the ego vehicle context features into the trajectory query. The transformer decoder employs a standard structure including multi-head self attention, multi-head cross attention and feed forward layers.
\begin{equation}
{Q}^{m+1}=\mathrm{TransDecoder}(q=s,k=\tilde{S^0},v=\tilde{S^0})
  \end{equation}
where \(\mathrm{TransDecoder}\) represents the transformer decoder, and  \(\tilde{S^0}\in\mathbb{R}^{c\times T_h\times D}\) is the ego vehicle context features copied \(c\) times.

In the final decoder layer, we attach an MLP to generate future trajectories from the trajectory query \(Q^M\). The output \(Y\in\mathbb{R}^{c\times T_f\times 7}\) includes the position \((\hat{\mu}_{x,t}^c,\hat{\mu}_{y,t}^c,\hat{\sigma}_{x,t}^c,\hat{\sigma}_{y,t}^c,\hat{\rho}_t^c)\) and the velocity \((\hat{v}_{x,t}^c,\hat{v}_{y,t}^c)\) of the ego vehicle for each mode and each time step. In addition, we apply another MLP followed by a softmax to \(Q^M\) to obtain the mode probabilities \(\hat{p}^c\).

\begin{table*}[t]
\caption{minADE, minFDE, and RMSE (m) of RHP and the benchmark models on highD and SHRP2 datasets. \textbf{Bold} is the best performance, and \underline{underline} is the second-best.}
  \label{tab:baselines}
  \begin{center}
    \begin{small}
  \begin{tabular}{lcccccc}
    \toprule
			\multirow{2}{*}{Dataset} &\multirow{2}{*}{Methods} &\multicolumn{5}{c}{Prediction Horizon (minADE\(\downarrow\) / minFDE\(\downarrow\) / RMSE\(\downarrow\))}\\
                \cmidrule(lr){3-7}
                 & & 1 & 2 & 3 & 4 & 5 \\
                \midrule
  \multirow{8}{*}{highD} &S-LSTM  & 0.19/0.20/0.26 & 0.26/0.44/0.60 & 0.39/0.83/1.16 & 0.58/1.34/1.91 & 0.81/1.98/2.86 \\
  &CS-LSTM  & 0.14/0.16/0.23 & 0.22/0.38/0.61 & 0.32/0.66/1.22 & 0.47/1.09/2.07 & 0.66/1.65/3.20 \\
  &WSiP  & 0.11/0.16/0.23 & 0.20/0.39/0.56 & 0.33/0.72/1.10 & 0.49/1.15/1.79 & 0.69/1.71/2.65 \\
  &GNP   & 0.21/0.31/0.46 & 0.30/0.43/0.92 & 0.35/0.44/1.56 & 0.38/0.56/2.42 & 0.46/0.98/3.50\\
  &PiP    & 0.12/0.18/0.17 & 0.22/0.43/0.55 & 0.35/0.75/0.98 & 0.51/1.16/1.53 & 0.70/1.71/2.27 \\
  &STDAN & 0.04/0.06/\underline{0.08} & 0.07/0.14/0.19 & 0.12/0.28/0.38 & 0.20/\underline{0.53}/0.73& 0.31/\underline{0.92}/1.33 \\
  &HLTP    & \underline{0.04}/\underline{0.05}/0.15 & \underline{0.06}/\underline{0.11}/\underline{0.17} & \underline{0.11}/\underline{0.27}/\underline{0.35} & \underline{0.19}/0.54/\underline{0.71} & \underline{0.30}/0.94/\underline{1.32} \\
  &\cellcolor{green!12}RHP & \cellcolor{green!12}\textbf{0.04/0.05/0.06} & \cellcolor{green!12}\textbf{0.05/0.06/0.08} & \cellcolor{green!12}\textbf{0.06/0.11/0.20} & \cellcolor{green!12}\textbf{0.09/0.21/0.51}& \cellcolor{green!12}\textbf{0.13/0.38/0.99}\\
                 \midrule
  \multirow{8}{*}{SHRP2} &S-LSTM   &0.96/0.84/1.26&1.08/1.54/2.37&1.49/2.90/4.29&2.13/4.84/6.87&2.95/7.24/10.17\\
  &CS-LSTM &0.40/0.60/0.84&0.74/1.45/2.06&1.23/2.77/3.89&1.88/4.62/6.57&2.72/7.28/10.57\\
  &WSiP    &\underline{0.21}/\underline{0.41}/0.66&\underline{0.54}/1.22/1.82&1.01/2.47/3.53&1.62/4.18/5.82&2.38/6.35/8.66\\
  &GNP*    &0.60/0.90/1.09&0.88/1.29/2.53&1.05/\underline{1.42}/4.65&\underline{1.20}/\underline{1.92}/7.49&\underline{1.50}/\underline{3.33}/10.97\\
  &PiP     &0.38/0.52/\underline{0.64}&0.59/1.02/1.72&0.89/1.85/3.30&1.32/3.17/5.41&1.90/5.04/\underline{8.05}\\
  &STDAN   &0.49/0.56/0.78&0.70/1.22/1.69&1.07/2.24/3.22&1.60/3.94/6.06&2.19/7.26/12.03\\
  &HLTP    &0.39/0.56/1.01&0.60/\underline{1.01}/\underline{1.51}&\underline{0.87}/1.72/\underline{2.55}&1.25/2.91/\underline{4.61}&1.82/5.11/8.47\\
  &\cellcolor{green!12}RHP &\cellcolor{green!12} \textbf{0.20/0.33/0.56} &\cellcolor{green!12} \textbf{0.37/0.69/1.51} & \cellcolor{green!12}\textbf{0.55/1.03/2.98} & \cellcolor{green!12}\textbf{0.73/1.46/4.98}& \cellcolor{green!12}\textbf{0.97/2.36/7.47}\\
 \bottomrule
  \end{tabular}
  \end{small}
  \end{center}
  \footnotesize\textit{Note:} *no lane information
  \vskip -0.1in
\end{table*}

\subsection{Loss Function}

For the auxiliary coarse trajectory prediction task, the loss \(\mathcal{L}_{\text{coarse}}\)is defined as the mean squared error (MSE) of the predicted trajectories and the ground truth, as given below.
\begin{equation}
\mathcal{L}_{\text{coarse}}=\frac{1}{N_v}\frac{1}{T_e}\sum_{i}^{N_v} \sum_{t}^{T_e}\mathcal{L}^{M}(\hat{G}^{i}_t, {G}_{t}^i)
\end{equation}
where \({G}_{t}^i\) represents the ground truth position and velocity vector of vehicle \(i\) at horizon endpoint \(t\), \(\hat{G}^{i}_t\) is the predicted coarse trajectory and velocity of vehicle \(i\), and \(\mathcal{L}^{M}\) is the MSE loss function.

For final trajectory prediction, the objective is to assign high probability to the best-matching candidate while minimizing its distance from the ground-truth trajectory. To this end, we use a negative log-likelihood (NLL) loss to fit the bivariate Gaussian of the best predicted trajectory and an MSE loss to regress its velocity. In addition, mode prediction over candidate probabilities is formulated as a classification problem, where cross-entropy (CE) loss is employed to maximize the likelihood of the selected mode. 
\begin{align}
&a=\arg\min_c \left(\frac{1}{T_f}\sum_{t = 1}^{T_f}||\hat{\mu}_{t}^{c} - u_{t}||_2\right)\\
&\mathcal{L}_{\text{traj}}=\frac{1}{T_f}\sum_{t = 1}^{T_f} \left(\mathcal{L}^{M}(\hat{v}_{t}^{a},v_{t})+\mathcal{L}^{N}(\hat{Y}_{t}^{a},Y_{t})\right)+\gamma_0 \mathcal{L}^{cls}
\end{align}
where \(\mathcal{L}^{N}\) is the NLL loss function, \(\mathcal{L}^{cls}\) is the cross-entropy function, \(a\) is the index of the best-matching trajectory, \(\hat{\mu}_{t}^{a}\) is the mean position of the best trajectory, \(u_t\) is the ground truth position, \(\hat{v}_{t}^{a}\) is the predicted velocity, \(v_{t}\) is the ground truth velocity, and \(\gamma_0\) is a constant scaling factor for cross-entropy loss.

The final loss function is the sum of \(\mathcal{L}_{\text{coarse}}\) and trajectory prediction loss \(\mathcal{L}_{\text{traj}}\), weighted by \(\gamma_1\) and \(\gamma_2\).
\begin{equation}
  \mathcal{L}=\gamma_1 \mathcal{L}_{\text{coarse}}+\gamma_2 \mathcal{L}_{\text{traj}}
\end{equation}

\section{Experiments}
In this section, we present both quantitative and qualitative results of our model’s performance.

\subsection{Experimental Setup}
\textbf{Datasets.} The datasets used in this study are selected to cover both highway and urban driving environments and to support the analysis of safety-critical scenarios. We use two datasets: highD (Highway Drone) \cite{krajewski2018highd} and bird's eye view reconstructed SHRP2 (Second Strategic Highway Research Program's Naturalistic Driving Study) dataset \cite{jiao2025shrpcrash}. HighD includes real highway driving scenarios collected from drone video recordings at 25 Hz. Although it contains fewer recorded scenarios than large-scale benchmarks such as WOMD and Argoverse, its long recording duration and dense multi-vehicle observations allow us to extract approximately 17 million trajectory samples through sliding-window processing, which is suitable for large-sample evaluation. SHRP2 records naturalistic urban driving behavior using in-vehicle sensors and cameras \cite{hankey2016description}, and Jiao and Calvert \cite{jiao2025shrpcrash} further reconstruct its trajectories into a bird’s-eye-view representation. Compared with highD, SHRP2 is much smaller in scale, but it contains many safety-critical interactions on urban streets, including trajectories associated with crash scenes. More importantly, it provides ground-truth labels for safe, near-crash, and crash scenarios, which are unavailable in highD, WOMD, and Argoverse. These characteristics make SHRP2 well suited for safety-oriented evaluation.

\begin{figure*}[!t]
\centering
\subfloat[Visualization of predictions on SHRP2.]{\includegraphics[width=\textwidth]{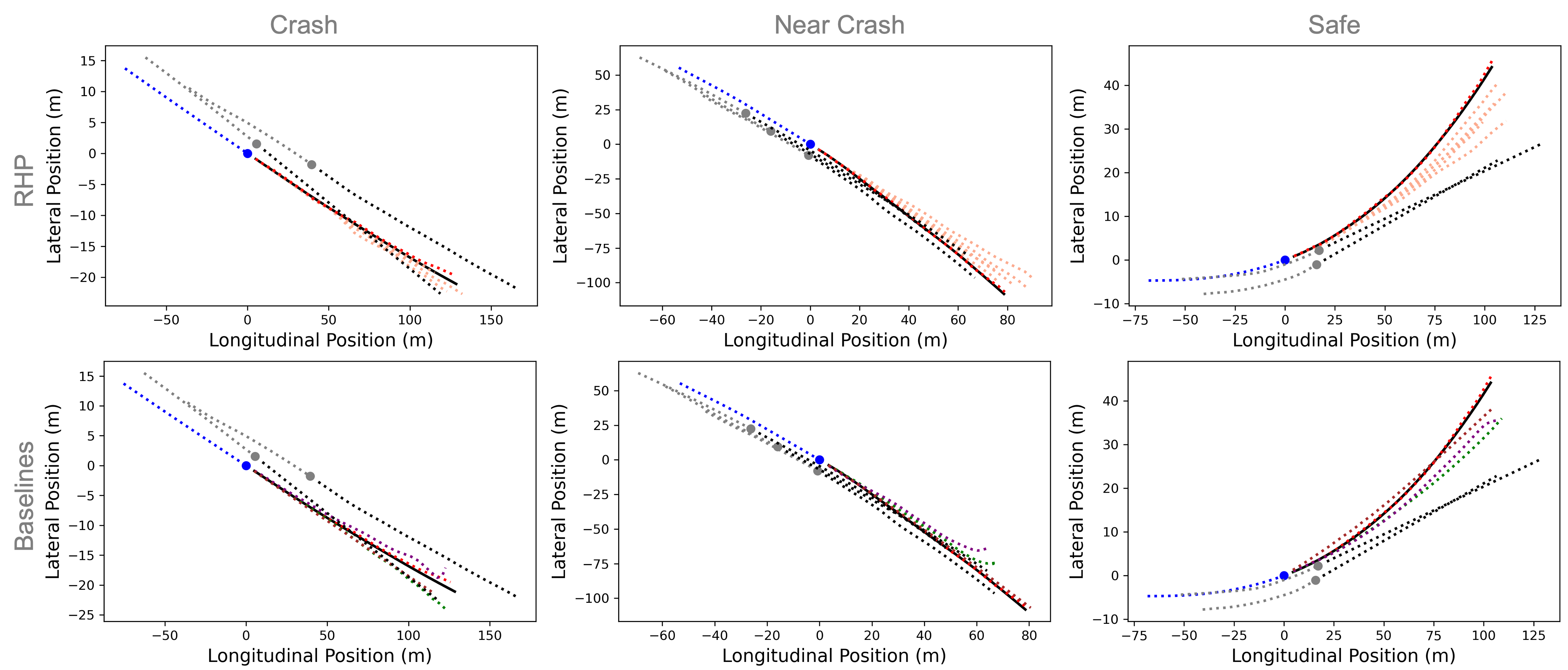}%
\label{fig:vis:sub1}}
\hfil
\subfloat[Visualization of predictions on highD.]{\includegraphics[width=\textwidth]{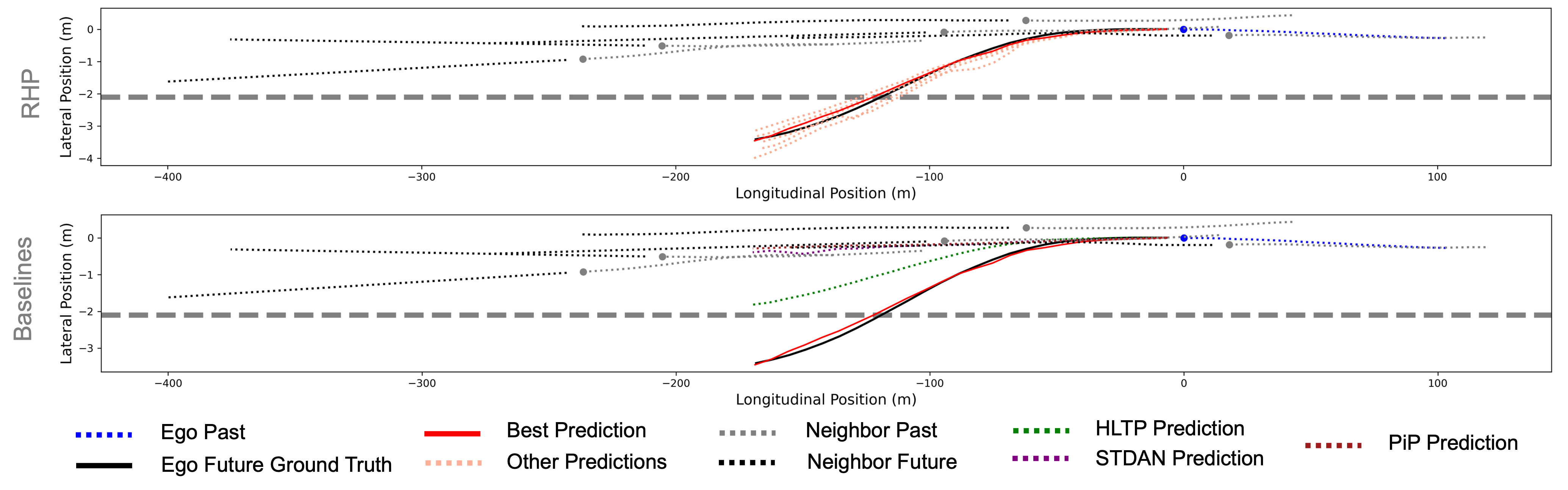}%
\label{fig:vis:sub2}}
\caption{Visualization of our RHP model with various models on SHRP2 and highD datasets. The top panel illustrates the multi-modal trajectory predictions of RHP, with the best prediction marked in \textcolor{red}{red} and other predictions marked in \textcolor{pink}{pink}. The bottom panel illustrates the best predictions of RHP with state-of-the-art baseline models.}
\label{fig:vis}
\end{figure*}

\textbf{Metrics.} We evaluate the model using average displacement error (ADE), final displacement error (FDE), and root mean squared error (RMSE). ADE calculates the average \(\ell_2\) distance between the predicted and actual positions over the entire trajectory. FDE calculates the \(\ell_2\) distance at the final timestep of the prediction. RMSE measures the square root of the mean of squared position errors across all samples. To evaluate the model’s ability to cover the ground-truth motion, we report the minimum ADE and FDE over the \(c\) predicted modes to measure best-case trajectory accuracy. For RMSE, we compute it on the single trajectory with the highest probability, which is the standard evaluation protocol used for the highD dataset. In addition, we examine the distribution of horizon-importance weights over 1–5 seconds to measure whether the model attends to appropriate future horizons under high-risk interactions.

\textbf{Implementation Details.} The experiment was conducted on a machine equipped with an NVIDIA GeForce RTX 4090 GPU and an Intel Core i9-14900KF CPU. For both datasets, the model predicts the next 1--5 seconds based on the past 3 seconds of observed trajectories. The hidden feature dimension of both the encoder and decoder is set to $D = 64$, with 3 encoder layers and 2 decoder layers used for SHRP2, and 1 encoder layer and 1 decoder layer used for highD. In the risk horizon profiling module, the number of intention trajectories is set to \(J=100\). The model is trained end-to-end for 12 epochs using the Adam optimizer with a batch size of 128. The initial learning rate is 0.0005 and is decayed by a factor of 0.4 after each epoch. For the final prediction, the model outputs \(c=6\) trajectory modes.

\subsection{Quantitative Results}
We compare our model against S-LSTM \cite{alahi2016social}, CS-LSTM \cite{deo2018convolutional}, WSiP \cite{wang2023wsip}, GNP \cite{gan2025goal}, PiP \cite{song2020pip}, STDAN \cite{chen2022intention}, and HLTP \cite{liao2024cognitive}. For GNP, we disable lane-dependent modules since SHRP2 does not provide lane information, which may degrade its performance. For a fair comparison, we re-train all baselines on both highD and SHRP2. As shown in Tab. \ref{tab:baselines}, our method outperforms all baselines. Compared with the best-performing baseline, our method reduces the RMSE of the highest-probability trajectory by 25.0\% on highD. On SHRP2, it reduces 1s minFDE by 19.5\% and 5s minFDE by 29.1\%. These results indicate strong performance for both short and long horizon prediction and robust generalization across highway and urban scenarios even without map priors.

\subsection{Qualitative Results}

We further assess prediction accuracy through qualitative analysis. Fig. \ref{fig:vis:sub1} visualizes crash, near-crash, and safe scenarios from SHRP2, comparing our multi-modal trajectory predictions with baselines including HLTP, STDAN, and PiP. Our method produces future trajectories that closely match the ground truth and exhibits smooth, adaptive behavior across crash, near-crash, and safe scenarios. Fig. \ref{fig:vis:sub2} presents a lane-change scenario from highD. Baseline models fail to capture the lane-change intent, often predicting straight trajectories or only slight deviations. RHP correctly anticipates the maneuver, demonstrating strong robustness and understanding of complex driving intentions.

\begin{figure}[ht]
  \begin{center}
    \centerline{\includegraphics[width=\columnwidth]{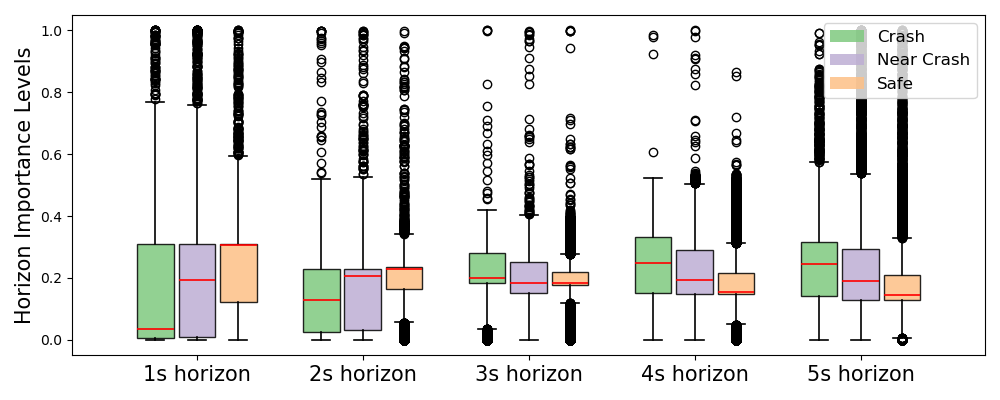}}
    \caption{Horizon importance levels under crash, near-crash, and safe scenarios in SHRP2 dataset.}
    \label{fig:attns}
  \end{center}
\end{figure}

To interpret the physical meaning of the horizon importance weights learned by the RHP module, we visualize the weight levels in Fig. \ref{fig:attns} over 1 to 5-second horizons using box plots on SHRP2 under crash, near-crash, and safe scenarios. Overall, crash cases show higher importance at longer horizons, with greater emphasis on the 5-second risk distribution, whereas safe cases place more weight on short-horizon risk, particularly at 1 second. This trend is consistent with our intuition that drivers rely more on reflexive, short-term cues in normal traffic, but shift attention toward longer-horizon risk patterns when the situation becomes hazardous and proactive anticipation is required. These results suggest that RHP captures the scenario-dependent uncertainty underlying human driving behavior, and that its adaptive horizon importance mitigates bias induced by scenario imbalance by dynamically reallocating attention across future horizons.

\begin{figure}[ht]
\begin{center}
\subfloat[Objective risk perceived by the  vehicle along time frames.]{\includegraphics[width=0.48\textwidth]{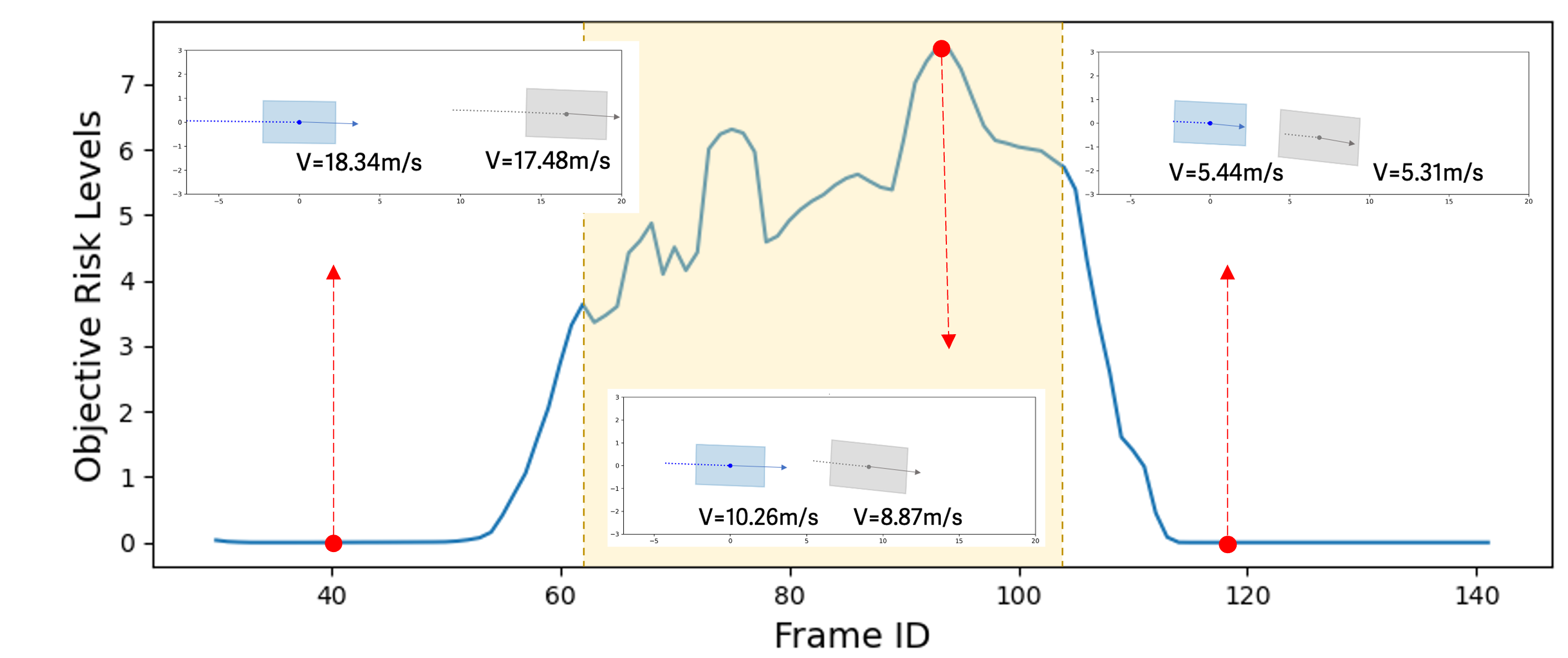}%
\label{fig:vis:sub1}}
\hfil
\subfloat[Inferred horizon importance levels along time frames.]{\includegraphics[width=0.48\textwidth]{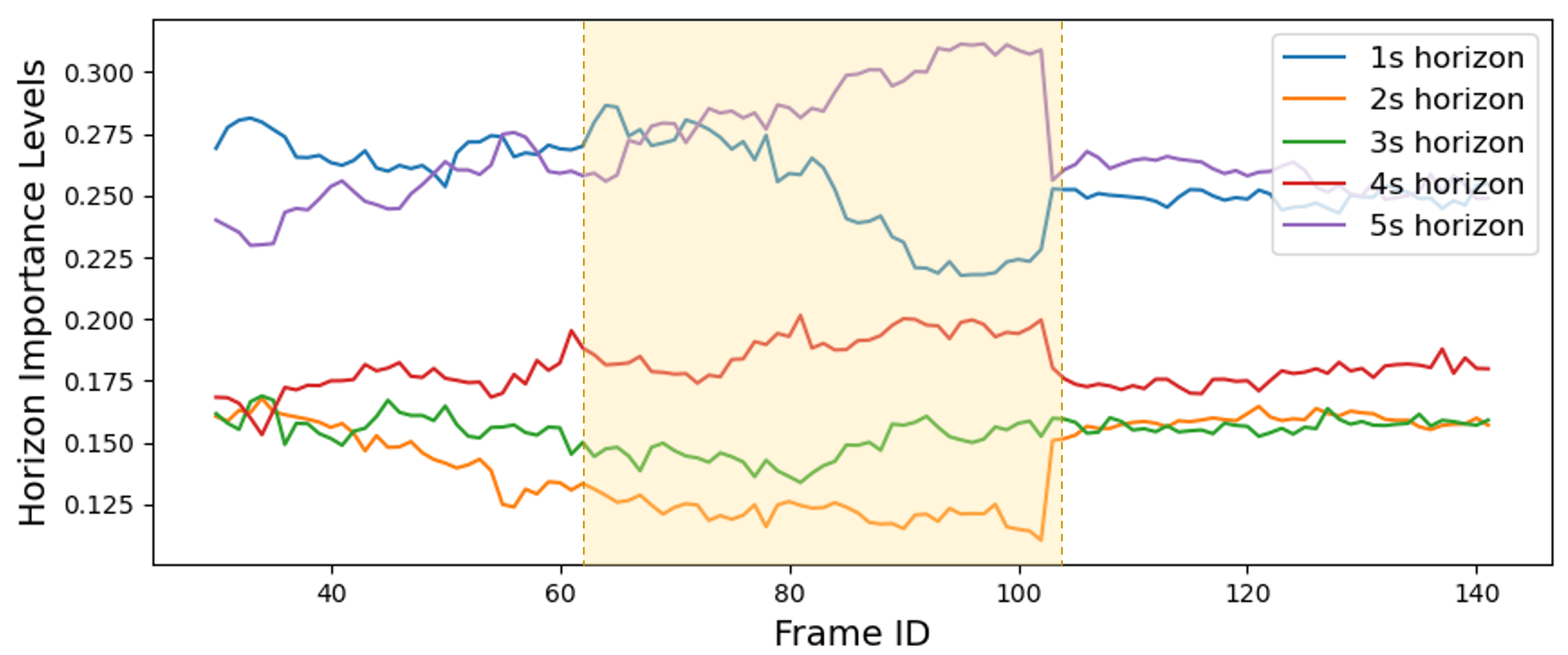}%
\label{fig:vis:sub2}}
\caption{Objective risk evolution and horizon importance dynamics in a SHRP2 near-crash scenario.}
\label{fig:attn_orisk}
\end{center}
\end{figure}

We further analyze a representative near-crash scenario from SHRP2, as shown in Fig. \ref{fig:attn_orisk}. The blue ego vehicle approaches a decelerating gray vehicle ahead. As their distance narrows, the objective risk increases, accompanied by a rise in the 5-second horizon importance and a decrease in the 1-second horizon importance, indicating the growing relevance of longer-term risk. As the ego vehicle slows down, the collision probability drops despite the short distance, and both the objective risk and the horizon importance levels return toward their initial states.

\subsection{Ablation Study}

\begin{table}[t]
\caption{
Ablation setting of RHP components. 
}
\begin{center}
\begin{small}
\begin{tabular}{lccc}
\toprule
& \multicolumn{3}{c}{Components} \\
\cmidrule(lr){2-4}
Method 
& Coarse endpoints 
& Endpoint risk 
& RHP query \\
\midrule
A 
& -- & -- & -- \\
B 
& \cmark & -- & -- \\
C
& \cmark & $R_5$ only & -- \\
D 
& \cmark & $R_{1:5}$ & -- \\
E 
& \cmark & $R_{1:5}$ & Learned query \\
F 
& \cmark & $R_{1:5}$ & Mean($R_{1:5}$) \\
G
& \cmark & $R_{1:5}$ & $R_5$ \\
H
& \cmark & $R_{1:5}$ & $R_3$ \\
Ours
& \cmark & $R_{1:5}$ & $R_1$ \\

\bottomrule
\end{tabular}
\label{tab:ablation_main_simplified}
\end{small}
\end{center}
\footnotesize\textit{Note:} \textbf{Coarse endpoints}: The auxiliary coarse endpoint trajectory prediction module. 
\textbf{Endpoint risk}: The horizon endpoint risk used in the RHP module. $R_{1:5}$ is the 1 to 5-second endpoint risk while $R_5$ is the 5-second endpoint risk. 
\textbf{RHP query}: The attention query used in the RHP module. 
\vskip -0.1in
\end{table}

\begin{table}[t]
\caption{Ablation results on SHRP2 across scenarios with different risk levels under a 5-second prediction horizon.}
\begin{center}
\begin{small}
\begin{tabular}{lcccccc}
\toprule
\multirow{3}{*}{Method} &\multicolumn{2}{l}{Collision-in-5s}&\multicolumn{2}{l}{Collision-in-10s} &\multicolumn{2}{l}{Non-collision} \\
&\multicolumn{2}{c}{(5.16\%)}&\multicolumn{2}{c}{(9.77\%)}&\multicolumn{2}{c}{(90.23\%)}\\
\cmidrule(lr){2-7}
& ADE & FDE& ADE & FDE & ADE & FDE \\
\midrule
 A & 1.18 & 2.67 & 1.09 & 2.55 & 1.00 & 2.45\\
 B & 1.15 & 2.57 & 1.07 & 2.46 & 0.98 & 2.40\\
 C & 1.15 & 2.56 & 1.06 & 2.46 & 0.98 & 2.41\\
 D & 1.17 & 2.61 & 1.09 & 2.51 & 1.00 & 2.44\\
 E & \underline{1.14} & \underline{2.55} & 1.06 & 2.45 & 0.98 & 2.40\\
 F & 1.14 & 2.56 & \underline{1.05} & \underline{2.44} & \underline{0.97} & \underline{2.37} \\
 G & 1.16 & 2.56 & 1.07 & 2.45 & 0.98 & 2.41\\
 H & 1.15 & 2.58 & 1.06 & 2.44 & 0.97 & 2.38 \\
 \cellcolor{green!12}Ours & \cellcolor{green!12}\textbf{1.13} & \cellcolor{green!12}\textbf{2.51} & \cellcolor{green!12}\textbf{1.05} & \cellcolor{green!12}\textbf{2.40} & \cellcolor{green!12}\textbf{0.96} & \cellcolor{green!12}\textbf{2.36}\\
\bottomrule
\end{tabular}
\label{tab:shrp_ab}
\end{small}
\end{center}
\vskip -0.1in
\end{table}

\begin{table}[t]
\caption{Ablation results on highD across scenarios with different risk levels under a 5-second prediction horizon.}
\begin{center}
\begin{small}
\begin{tabular}{lcccccc}
\toprule
\multirow{3}{*}{Method} &\multicolumn{2}{l}{Collision-in-5s}&\multicolumn{2}{l}{Collision-in-10s} &\multicolumn{2}{l}{Non-collision} \\
&\multicolumn{2}{c}{(0.16\%)}&\multicolumn{2}{c}{(2.51\%)}&\multicolumn{2}{c}{(97.49\%)}\\
\cmidrule(lr){2-7}
& ADE & FDE& ADE & FDE & ADE & FDE \\
\midrule
 A & 0.28 & 0.89 & 0.21 & 0.65 & 0.15 & 0.44\\
 B & {0.30} & {0.80} & {0.26} & {0.74} & {0.15} & {0.45}\\
 C & 0.22 & 0.62 & 0.20 & 0.62 & 0.15 & 0.46\\
 D & 0.28&0.77&0.23&0.62&\underline{0.13}&\underline{0.37}\\
 E & 0.23&0.73&0.21&0.67&0.15&0.47\\
 F & 0.23&0.64&0.19&0.56&0.14&0.40\\
 G & 0.23&\underline{0.62}&0.21&0.60&0.14&0.42\\
 H & \underline{0.22}&0.67&\underline{0.18}&\textbf{0.53}&0.13&0.39\\
 \cellcolor{green!12}Ours & \cellcolor{green!12}\textbf{0.22} & \cellcolor{green!12}\textbf{0.62} & \cellcolor{green!12}\textbf{0.18} & \cellcolor{green!12}\underline{0.54} & \cellcolor{green!12}\textbf{0.13} & \cellcolor{green!12}\textbf{0.37}\\
\bottomrule
\end{tabular}
\label{tab:highd_ab}
\end{small}
\end{center}
\vskip -0.1in
\end{table}

To quantify the contribution of each component to overall performance, we conduct a series of ablation studies, as reported in Tab. \ref{tab:ablation_main_simplified}, \ref{tab:shrp_ab}, \ref{tab:highd_ab}. Introducing the auxiliary coarse endpoint prediction task (B vs. A) reduces errors across all scenarios on SHRP2, but yields only limited gains on highD. This suggests that lightweight horizon endpoint estimates provide useful structural cues for downstream prediction, but are not sufficient to ensure robust improvements across all datasets. Adding only the final 5-second risk (C) improves high-risk performance on highD, but the gains are not consistent across all scenarios. Feeding all-horizon risks (D) into the decoder without modeling horizon-wise importance improves non-collision performance on highD, but performs poorly in the collision-in-5s group and is worse than Method B across all metrics on SHRP2. This suggests that uniformly treating multi-horizon risk cues may dilute decision-critical signals, highlighting the necessity of risk horizon profiling

Methods E–H and our final model all adopt risk horizon profiling, but differ in the query design used in the RHP module. The results show that learned (E), averaged (F), long-horizon (G), and mid-horizon risk queries (H) can achieve competitive performance in some cases. However, using 1-second endpoint risk as the query in our final model leads to the best overall results, suggesting that short-horizon risk provides a more reliable and controllable reference for reweighting longer-horizon and potentially noisier risk patterns. Notably, our full model yields the largest improvements in collision-prone scenarios, demonstrating stronger robustness and generalization under high-risk interactions.

\begin{table}[t]
\centering
\caption{Ablation study on endpoint prediction and future risk contribution.}
\label{tab:endpoint_risk_ablation}
\begin{tabular}{lllcc}
\hline
Method & Endpoint prediction & Endpoint risk & ADE & FDE \\
\hline
I & -- & -- & 1.03 & 2.51 \\
II & -- & Constant Velocity & 1.05 & 2.56 \\
III & -- & Ground-Truth & \textbf{0.91} & \textbf{2.16} \\
\cellcolor{green!12}Ours & \cellcolor{green!12}\cmark & \cellcolor{green!12}RHP & \cellcolor{green!12}\underline{0.96} & \cellcolor{green!12}\underline{2.36} \\
\hline
\end{tabular}
\end{table}

To further separate the contributions of endpoint prediction and risk horizon modeling, we conduct an additional ablation study with four configurations, as shown in Tab. \ref{tab:endpoint_risk_ablation}. Method I removes both endpoint prediction and risk modeling, serving as a risk-free baseline. Method II replaces the learned endpoint predictions with constant-velocity kinematic extrapolation to estimate future vehicle positions, which are then used to compute the risk profile. Due to its simplified motion assumption, this configuration may introduce greater noise into risk estimation. Method III computes the risk profile using ground-truth future positions. Although this setting introduces data leakage and is therefore not a valid deployable model, it provides an informative upper bound by removing endpoint prediction errors from the risk estimation process.

The results indicate that future risk modeling provides gains beyond the contribution of endpoint prediction. However, the constant-velocity risk variant performs worse than the risk-free baseline, suggesting that inaccurate future position estimates can produce noisy risk profiles and mislead the horizon-importance attention mechanism. When the risk profile is computed from ground-truth future positions, the model achieves the best performance, further confirming that more accurately calibrated future risk estimates can improve trajectory prediction.


\subsection{Inference Latency} 
The RHP model achieves an average inference latency of 13 ms with 1 encoder layer, 1 decoder layer, and a batch size of 1, measured using standard PyTorch implementation on a NVIDIA RTX 4090 GPU. During each inference, the model predicts six multimodal future trajectories for the ego vehicle while also estimating coarse endpoints for all agents in the scene, with up to 30 agents considered.

\section{Conclusion}
In this paper, we present a risk-evolution-aware trajectory prediction framework based on dynamic risk horizon profiling. We quantify risk using a continuous, learnable risk potential field and model the evolution and uncertainty of future risk across candidate trajectories and prediction horizons. We further introduce a risk horizon profiling module that learns horizon importance weights to adaptively highlight critical future moments under different risk scenarios. Compared with state-of-the-art prediction methods, our model achieves a 25.0\% reduction in RMSE on highD and a 29.1\% reduction in minFDE on SHRP2 over a 5-second prediction horizon, demonstrating strong robustness and generalization across safe, near-crash, and crash events.

Future work will extend the proposed framework from vehicle-to-vehicle interactions to heterogeneous traffic participants, including pedestrians and cyclists. This extension will require risk potential field models tailored to the motion patterns and interaction characteristics of different road users. The proposed framework will be further developed into a plug-and-play structure, providing a flexible foundation for integrating such agent-specific risk models. In addition, we will also incorporate additional safety-critical factors, including traffic rule compliance and road geometry, to enable more comprehensive risk assessment in real-world scenarios.


\section*{Acknowledgments}
This work was supported by the Connected Cities for Smart Mobility towards Accessible and Resilient Transportation (C2SMART) Center, a Tier 1 University Center awarded by U.S. Department of Transportation under the University Transportation Centers Program. The contents of this paper only reflect the views of the authors who are responsible for the facts and do not represent any official views of any sponsoring organizations or agencies.


\bibliographystyle{IEEEtran}
\bibliography{example_paper}

@article{liao2024cognitive,
  title={A cognitive-based trajectory prediction approach for autonomous driving},
  author={Liao, Haicheng and Li, Yongkang and Li, Zhenning and Wang, Chengyue and Cui, Zhiyong and Li, Shengbo Eben and Xu, Chengzhong},
  journal={IEEE Transactions on Intelligent Vehicles},
  year={2024},
  publisher={IEEE}
}

@article{li2024context,
  title={Context-aware trajectory prediction for autonomous driving in heterogeneous environments},
  author={Li, Zhenning and Chen, Zhiwei and Li, Yunjian and Xu, Chengzhong},
  journal={Computer-Aided Civil and Infrastructure Engineering},
  volume={39},
  number={1},
  pages={120--135},
  year={2024},
  publisher={Wiley Online Library}
}

@inproceedings{liang2020learning,
  title={Learning lane graph representations for motion forecasting},
  author={Liang, Ming and Yang, Bin and Hu, Rui and Chen, Yun and Liao, Renjie and Feng, Song and Urtasun, Raquel},
  booktitle={European Conference on Computer Vision},
  pages={541--556},
  year={2020},
  organization={Springer}
}

@article{weiss2026barte,
  title={Bart{\'e}: Composite b{\'e}zier curves for racing trajectory estimation},
  author={Weiss, Trent and Behl, Madhur},
  journal={Journal of Intelligent \& Robotic Systems},
  volume={112},
  number={2},
  pages={48},
  year={2026},
  publisher={Springer}
}

@article{huang2025post,
  title={Post-interactive Multimodal Trajectory Prediction for Autonomous Driving},
  author={Huang, Ziyi and Li, Yang and Li, Dushuai and Mu, Yao and Qin, Hongmao and Zheng, Nan},
  journal={arXiv preprint arXiv:2503.09366},
  year={2025}
}

@article{bharilya2024machine,
  title={Machine learning for autonomous vehicle's trajectory prediction: A comprehensive survey, challenges, and future research directions},
  author={Bharilya, Vibha and Kumar, Neetesh},
  journal={Vehicular Communications},
  volume={46},
  pages={100733},
  year={2024},
  publisher={Elsevier}
}

@article{feng2025risk,
  title={Risk-Aware Stochastic Vehicle Trajectory Prediction With Spatial-Temporal Interaction Modelling},
  author={Feng, Yuxiang and Ye, Qiming and Candela, Eduardo and Escribano-Macias, Jose and Hu, Bo and Demiris, Yiannis and Angeloudis, Panagiotis},
  journal={IEEE Open Journal of Intelligent Transportation Systems},
  year={2025},
  publisher={IEEE}
}

@article{wang2025risk,
  title={Risk-Aware Vehicle Trajectory Prediction Under Safety-Critical Scenarios},
  author={Wang, Qingfan and Xu, Dongyang and Kuang, Gaoyuan and Lv, Chen and Li, Shengbo Eben and Nie, Bingbing},
  journal={IEEE Transactions on Intelligent Transportation Systems},
  year={2025},
  publisher={IEEE}
}

@inproceedings{dang2023ttc,
  title={TTC-SLSTM: Human trajectory prediction using time-to-collision interaction energy},
  author={Dang, Huu-Tu and Korbmacher, Raphael and Tordeux, Antoine and Gaudou, Benoit and Verstaevel, Nicolas},
  booktitle={2023 15th International Conference on Knowledge and Systems Engineering (KSE)},
  pages={1--6},
  year={2023},
  organization={IEEE}
}

@techreport{hankey2016description,
  title={Description of the SHRP 2 naturalistic database and the crash, near-crash, and baseline data sets},
  author={Hankey, Jonathan M and Perez, Miguel A and McClafferty, Julie A},
  year={2016},
  institution={Virginia Tech Transportation Institute}
}

@inproceedings{song2020pip,
  title={Pip: Planning-informed trajectory prediction for autonomous driving},
  author={Song, Haoran and Ding, Wenchao and Chen, Yuxuan and Shen, Shaojie and Wang, Michael Yu and Chen, Qifeng},
  booktitle={European conference on computer vision},
  pages={598--614},
  year={2020},
  organization={Springer}
}

@inproceedings{thuremella2024risk,
  title={Risk-aware Trajectory Prediction by Incorporating Spatio-temporal Traffic Interaction Analysis},
  author={Thuremella, Divya and Ince, Lewis and Kunze, Lars},
  booktitle={2024 IEEE International Conference on Robotics and Automation (ICRA)},
  pages={14421--14427},
  year={2024},
  organization={IEEE}
}

@article{ning2025strap,
  title={STRAP: Spatial-Temporal Risk-Attentive Vehicle Trajectory Prediction for Autonomous Driving},
  author={Ning, Xinyi and Bian, Zilin and Zuo, Dachuan and Ergan, Semiha},
  journal={arXiv preprint arXiv:2507.08563},
  year={2025}
}

@inproceedings{wang2023wsip,
  title={Wsip: Wave superposition inspired pooling for dynamic interactions-aware trajectory prediction},
  author={Wang, Renzhi and Wang, Senzhang and Yan, Hao and Wang, Xiang},
  booktitle={Proceedings of the AAAI Conference on Artificial Intelligence},
  volume={37},
  number={4},
  pages={4685--4692},
  year={2023}
}

@report{jiao2025shrpcrash,
author = {Jiao, Yiru and Calvert, Simeon},
publisher = {VTTI},
title = {Bird’s eye view trajectory reconstruction of naturalistic crashes and near-crashes in the SHRP2 NDS},
year = {2025},
version = {V1},
doi = {10.15787/VTT1/EFYEJR},
}

@article{liao2025sa,
  title={SA-TP $^{2} $: A Safety-Aware Trajectory Prediction and Planning Model for Autonomous Driving},
  author={Liao, Haicheng and Li, Zhenning and Zhu, Kaiqun and Li, Keqiang and Xu, Chengzhong},
  journal={IEEE Transactions on Robotics},
  year={2025},
  publisher={IEEE}
}

@article{lin2021vehicle,
  title={Vehicle trajectory prediction using LSTMs with spatial--temporal attention mechanisms},
  author={Lin, Lei and Li, Weizi and Bi, Huikun and Qin, Lingqiao},
  journal={IEEE Intelligent Transportation Systems Magazine},
  volume={14},
  number={2},
  pages={197--208},
  year={2021},
  publisher={IEEE}
}

@article{zhang2022ai,
  title={AI-TP: Attention-based interaction-aware trajectory prediction for autonomous driving},
  author={Zhang, Kunpeng and Zhao, Liang and Dong, Chengxiang and Wu, Lan and Zheng, Liang},
  journal={IEEE Transactions on Intelligent Vehicles},
  volume={8},
  number={1},
  pages={73--83},
  year={2022},
  publisher={IEEE}
}

@article{gao2025ise,
  title={ISE-GT: Interaction strength-enhanced graph Transformer for explainable multi-agent trajectory prediction},
  author={Gao, Zhenhai and Liu, Dayu and Hu, Hongyu and Gao, Fei and Zhao, Rui and Yu, Tong},
  journal={Transportation Research Part C: Emerging Technologies},
  volume={179},
  pages={105296},
  year={2025},
  publisher={Elsevier}
}

@inproceedings{gu2021densetnt,
  title={Densetnt: End-to-end trajectory prediction from dense goal sets},
  author={Gu, Junru and Sun, Chen and Zhao, Hang},
  booktitle={Proceedings of the IEEE/CVF international conference on computer vision},
  pages={15303--15312},
  year={2021}
}

@inproceedings{chib2024ms,
  title={MS-TIP: imputation aware pedestrian trajectory prediction},
  author={Chib, Pranav Singh and Nath, Achintya and Kabra, Paritosh and Gupta, Ishu and Singh, Pravendra},
  booktitle={International Conference on Machine Learning},
  pages={8389--8402},
  year={2024},
  organization={PMLR}
}

@article{cao2025fif,
  title={FIF: future interaction forecasted for multi-agent trajectory prediction},
  author={Cao, Yang and Li, Peiqing and Ling, Xiao and Li, Qipeng},
  journal={Transportation Research Part C: Emerging Technologies},
  volume={177},
  pages={105190},
  year={2025},
  publisher={Elsevier}
}

@article{zhu2025incorporating,
  title={Incorporating safety field theory into interactive trajectory prediction between VRU and vehicle: An integrated spatial--temporal and risk-aware model},
  author={Zhu, Dianchen and Fan, Zheyan and Ma, Wei and Zhang, Xuxin and Chan, Ho-Yin and Zhao, Mingming},
  journal={Traffic Injury Prevention},
  volume={26},
  number={5},
  pages={567--576},
  year={2025},
  publisher={Taylor \& Francis}
}

@article{zhu2022interaction,
  title={Interaction-aware cut-in trajectory prediction and risk assessment in mixed traffic},
  author={Zhu, Xianglei and Hu, Wen and Deng, Zejian and Zhang, Jinwei and Hu, Fengqing and Zhou, Rui and Li, Keqiu and Wang, Fei-Yue},
  journal={IEEE/CAA Journal of Automatica Sinica},
  volume={9},
  number={10},
  pages={1752--1762},
  year={2022},
  publisher={IEEE}
}

@article{liu2022interactive,
  title={Interactive trajectory prediction using a driving risk map-integrated deep learning method for surrounding vehicles on highways},
  author={Liu, Xulei and Wang, Yafei and Jiang, Kun and Zhou, Zhisong and Nam, Kanghyun and Yin, Chengliang},
  journal={IEEE Transactions on Intelligent Transportation Systems},
  volume={23},
  number={10},
  pages={19076--19087},
  year={2022},
  publisher={IEEE}
}

@article{zuo2025composite,
  title={Composite Safety Potential Field for Highway Driving Risk Assessment},
  author={Zuo, Dachuan and Bian, Zilin and Zuo, Fan and Ozbay, Kaan},
  journal={arXiv preprint arXiv:2504.21158},
  year={2025}
}

@inproceedings{krajewski2018highd,
  title={The highd dataset: A drone dataset of naturalistic vehicle trajectories on german highways for validation of highly automated driving systems},
  author={Krajewski, Robert and Bock, Julian and Kloeker, Laurent and Eckstein, Lutz},
  booktitle={2018 21st international conference on intelligent transportation systems (ITSC)},
  pages={2118--2125},
  year={2018},
  organization={IEEE}
}

@inproceedings{alahi2016social,
  title={Social lstm: Human trajectory prediction in crowded spaces},
  author={Alahi, Alexandre and Goel, Kratarth and Ramanathan, Vignesh and Robicquet, Alexandre and Fei-Fei, Li and Savarese, Silvio},
  booktitle={Proceedings of the IEEE conference on computer vision and pattern recognition},
  pages={961--971},
  year={2016}
}

@inproceedings{deo2018convolutional,
  title={Convolutional social pooling for vehicle trajectory prediction},
  author={Deo, Nachiket and Trivedi, Mohan M},
  booktitle={Proceedings of the IEEE conference on computer vision and pattern recognition workshops},
  pages={1468--1476},
  year={2018}
}

@article{chen2022intention,
  title={Intention-aware vehicle trajectory prediction based on spatial-temporal dynamic attention network for internet of vehicles},
  author={Chen, Xiaobo and Zhang, Huanjia and Zhao, Feng and Hu, Yu and Tan, Chenkai and Yang, Jian},
  journal={IEEE Transactions on Intelligent Transportation Systems},
  volume={23},
  number={10},
  pages={19471--19483},
  year={2022},
  publisher={IEEE}
}

@article{wei2024intention,
  title={Intention-based and Risk-Aware Trajectory Prediction for Autonomous Driving in Complex Traffic Scenarios},
  author={Wei, Wen and Wang, Jiankun},
  journal={arXiv preprint arXiv:2409.15821},
  year={2024}
}

@article{zhao2020novel,
  title={A novel generation-adversarial-network-based vehicle trajectory prediction method for intelligent vehicular networks},
  author={Zhao, Liang and Liu, Yufei and Al-Dubai, Ahmed Y and Zomaya, Albert Y and Min, Geyong and Hawbani, Ammar},
  journal={IEEE Internet of Things Journal},
  volume={8},
  number={3},
  pages={2066--2077},
  year={2020},
  publisher={IEEE}
}

@article{fang2023heterogeneous,
  title={Heterogeneous trajectory forecasting via risk and scene graph learning},
  author={Fang, Jianwu and Zhu, Chen and Zhang, Pu and Yu, Hongkai and Xue, Jianru},
  journal={IEEE Transactions on Intelligent Transportation Systems},
  volume={24},
  number={11},
  pages={12078--12091},
  year={2023},
  publisher={IEEE}
}

@inproceedings{kim2017probabilistic,
  title={Probabilistic vehicle trajectory prediction over occupancy grid map via recurrent neural network},
  author={Kim, ByeoungDo and Kang, Chang Mook and Kim, Jaekyum and Lee, Seung Hi and Chung, Chung Choo and Choi, Jun Won},
  booktitle={2017 IEEE 20Th international conference on intelligent transportation systems (ITSC)},
  pages={399--404},
  year={2017},
  organization={IEEE}
}

@inproceedings{liao2024bat,
  title={Bat: Behavior-aware human-like trajectory prediction for autonomous driving},
  author={Liao, Haicheng and Li, Zhenning and Shen, Huanming and Zeng, Wenxuan and Liao, Dongping and Li, Guofa and Xu, Chengzhong},
  booktitle={Proceedings of the AAAI Conference on Artificial Intelligence},
  volume={38},
  number={9},
  pages={10332--10340},
  year={2024}
}

@article{gao2023dual,
  title={Dual transformer based prediction for lane change intentions and trajectories in mixed traffic environment},
  author={Gao, Kai and Li, Xunhao and Chen, Bin and Hu, Lin and Liu, Jian and Du, Ronghua and Li, Yongfu},
  journal={IEEE Transactions on Intelligent Transportation Systems},
  volume={24},
  number={6},
  pages={6203--6216},
  year={2023},
  publisher={IEEE}
}

@article{li2019grip++,
  title={Grip++: Enhanced graph-based interaction-aware trajectory prediction for autonomous driving},
  author={Li, Xin and Ying, Xiaowen and Chuah, Mooi Choo},
  journal={arXiv preprint arXiv:1907.07792},
  year={2019}
}

@article{shi2022motion,
  title={Motion transformer with global intention localization and local movement refinement},
  author={Shi, Shaoshuai and Jiang, Li and Dai, Dengxin and Schiele, Bernt},
  journal={Advances in Neural Information Processing Systems},
  volume={35},
  pages={6531--6543},
  year={2022}
}

@article{chen2023goal,
  title={Goal-guided and interaction-aware state refinement graph attention network for multi-agent trajectory prediction},
  author={Chen, Xiaobo and Luo, Fengbo and Zhao, Feng and Ye, Qiaolin},
  journal={IEEE Robotics and Automation Letters},
  volume={9},
  number={1},
  pages={57--64},
  year={2023},
  publisher={IEEE}
}

@inproceedings{zhao2021tnt,
  title={Tnt: Target-driven trajectory prediction},
  author={Zhao, Hang and Gao, Jiyang and Lan, Tian and Sun, Chen and Sapp, Ben and Varadarajan, Balakrishnan and Shen, Yue and Shen, Yi and Chai, Yuning and Schmid, Cordelia and others},
  booktitle={Conference on Robot Learning},
  pages={895--904},
  year={2021},
  organization={PMLR}
}

@article{gan2025goal,
  title={Goal-based neural physics vehicle trajectory prediction model},
  author={Gan, Rui and Shi, Haotian and Li, Pei and Wu, Keshu and An, Bocheng and You, Junwei and Li, Linheng and Ma, Junyi and Ma, Chengyuan and Ran, Bin},
  journal={Transportation Research Part C: Emerging Technologies},
  volume={179},
  pages={105283},
  year={2025},
  publisher={Elsevier}
}

@article{westny2024diffusion,
  title={Diffusion-based environment-aware trajectory prediction},
  author={Westny, Theodor and Olofsson, Bj{\"o}rn and Frisk, Erik},
  journal={arXiv preprint arXiv:2403.11643},
  year={2024}
}

@inproceedings{gupta2018social,
  title={Social gan: Socially acceptable trajectories with generative adversarial networks},
  author={Gupta, Agrim and Johnson, Justin and Fei-Fei, Li and Savarese, Silvio and Alahi, Alexandre},
  booktitle={Proceedings of the IEEE conference on computer vision and pattern recognition},
  pages={2255--2264},
  year={2018}
}

@article{girgis2021latent,
  title={Latent variable sequential set transformers for joint multi-agent motion prediction},
  author={Girgis, Roger and Golemo, Florian and Codevilla, Felipe and Weiss, Martin and D'Souza, Jim Aldon and Kahou, Samira Ebrahimi and Heide, Felix and Pal, Christopher},
  journal={arXiv preprint arXiv:2104.00563},
  year={2021}
}

@inproceedings{yang2024intp,
  title={INTP-DM: Intention-Aware Multimodal Vehicle Trajectory Prediction Using Diffusion Model},
  author={Yang, Yuanbo and Jin, Sheng and Na, Xiaoxiang and Angeloudis, Panagiotis and Hu, Simon},
  booktitle={2024 IEEE 27th International Conference on Intelligent Transportation Systems (ITSC)},
  pages={3725--3730},
  year={2024},
  organization={IEEE}
}

@article{li2020evolvegraph,
  title={Evolvegraph: Multi-agent trajectory prediction with dynamic relational reasoning},
  author={Li, Jiachen and Yang, Fan and Tomizuka, Masayoshi and Choi, Chiho},
  journal={Advances in neural information processing systems},
  volume={33},
  pages={19783--19794},
  year={2020}
}

@article{chen2022vehicle,
  title={Vehicle trajectory prediction based on intention-aware non-autoregressive transformer with multi-attention learning for Internet of Vehicles},
  author={Chen, Xiaobo and Zhang, Huanjia and Zhao, Feng and Cai, Yingfeng and Wang, Hai and Ye, Qiaolin},
  journal={IEEE Transactions on Instrumentation and Measurement},
  volume={71},
  pages={1--12},
  year={2022},
  publisher={IEEE}
}

@article{zhu2026scenefactory,
  title={SceneFactory: GPU-Accelerated Multi-Agent Driving Simulation with Physics-Based Vehicle Dynamics},
  author={Zhu, Yicheng and Chen, Yang and Li, Tao and Bian, Zilin},
  journal={arXiv preprint arXiv:2605.08528},
  year={2026}
}

\newpage

\begin{IEEEbiography}[{\includegraphics[width=1in,height=1.1in,clip,keepaspectratio]{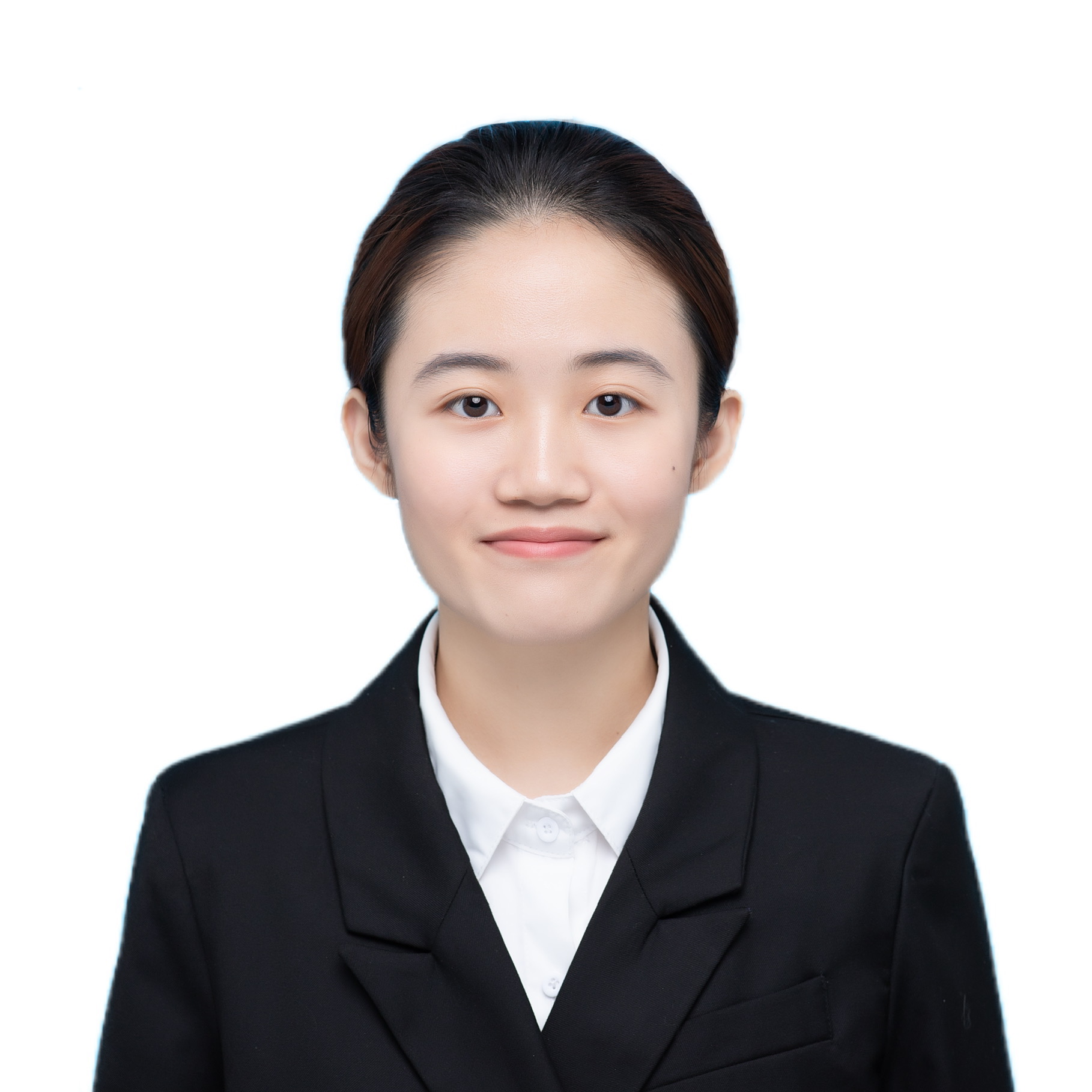}}]{Xinyi Ning} is a PhD candidate in Urban Systems at New York University Tandon School of Engineering, starting in the Fall of 2024. She earned her B.Sc. degree in Automation from Tsinghua University in 2021 and her M.Sc. degree in Control Science and Engineering from Tsinghua University in 2024. Her research focuses on intelligent transportation systems, autonomous driving safety, and transportation scene understanding. Her work also covers vehicle trajectory prediction and planning, traffic risk estimation, and driving behavior modeling.
\end{IEEEbiography}

\begin{IEEEbiography}
[{\includegraphics[width=1in,height=1.1in,clip,keepaspectratio]{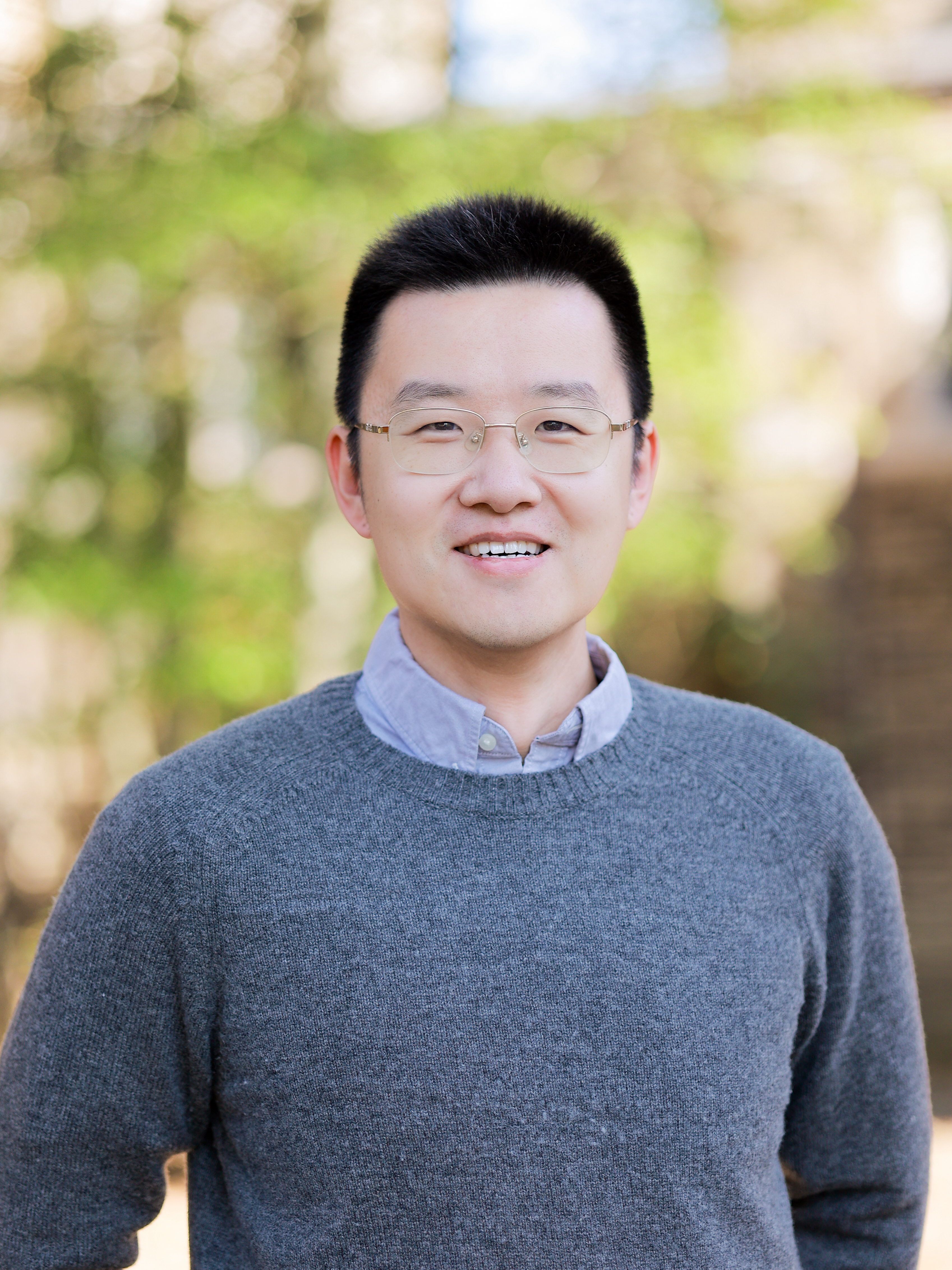}}]{Zilin Bian}~(Member, IEEE) is an Assistant Professor in the Department of Civil Engineering Technology and Environmental Safety Management at Rochester Institute of Technology. He holds a Ph.D. in Transportation Planning and Engineering from NYU, an M.S. in Civil Engineering from the University of Florida, and a B.S. from Harbin Institute of Technology. His research focuses on human-centered urban management decision science, smart and connected infrastructure systems, and cooperative digital twin technologies, aiming to enhance mobility, safety, resilience, and sustainability in urban transportation. Zilin has published 30+ refereed papers in peer-reviewed journals and conference proceedings. He also served as PI/Co-PI in securing research funding from state agencies and industry partners (e.g., NVIDIA). He was one of the recipients of the 2023 Best Dissertation Award from the TRB Artificial Intelligence Committee.
\end{IEEEbiography}

\begin{IEEEbiography}
[{\includegraphics[width=1in,height=1.1in,clip,keepaspectratio]{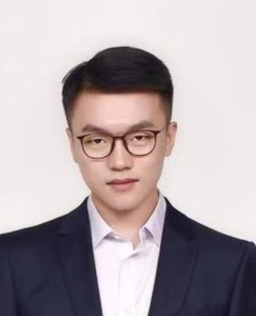}}]{Dachuan Zuo} is a Ph.D. candidate in Transportation and Highway Engineering at New York University Tandon School of Engineering. He received his M.S. from Northwestern University and his B.S. from Southwest Jiaotong University. His research focuses on data-driven safety modeling for naturalistic driving and autonomous vehicles, as well as safety-informed driving behavior modeling. His work also includes imitation learning for driving behavior generation in safety-critical scenarios, surrogate safety measures for roadway segments, and graph neural networks for large-scale traffic forecasting.
\end{IEEEbiography}

\begin{IEEEbiography}
[{\includegraphics[width=1in,height=1.1in,clip,keepaspectratio]{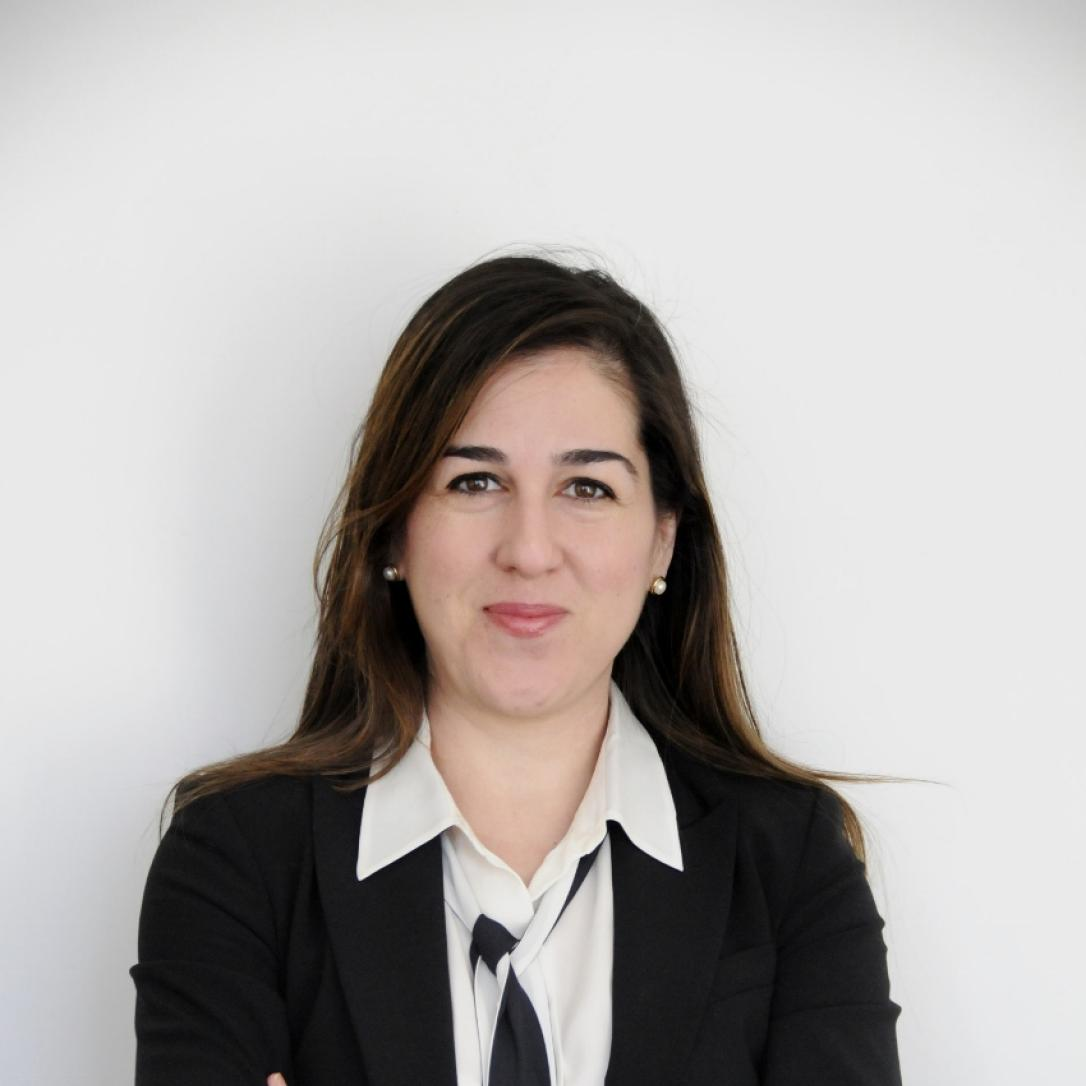}}]{Semiha Ergan} is an Associate Professor and Associate Department Chair in the Department of Civil and Urban Engineering at New York University. She received her Ph.D. in Computer-Aided Engineering from Carnegie Mellon University. Her research interests include augmented and virtual reality, autonomous driving, building information modeling, digital twins, computer vision, machine learning, data analytics, and visualization. Her work also explores building and urban informatics to support smarter, safer, and more efficient design, construction, operation, and management of buildings and urban systems.
\end{IEEEbiography}

\begin{IEEEbiography}
[{\includegraphics[width=1in,height=1.1in,clip,keepaspectratio]{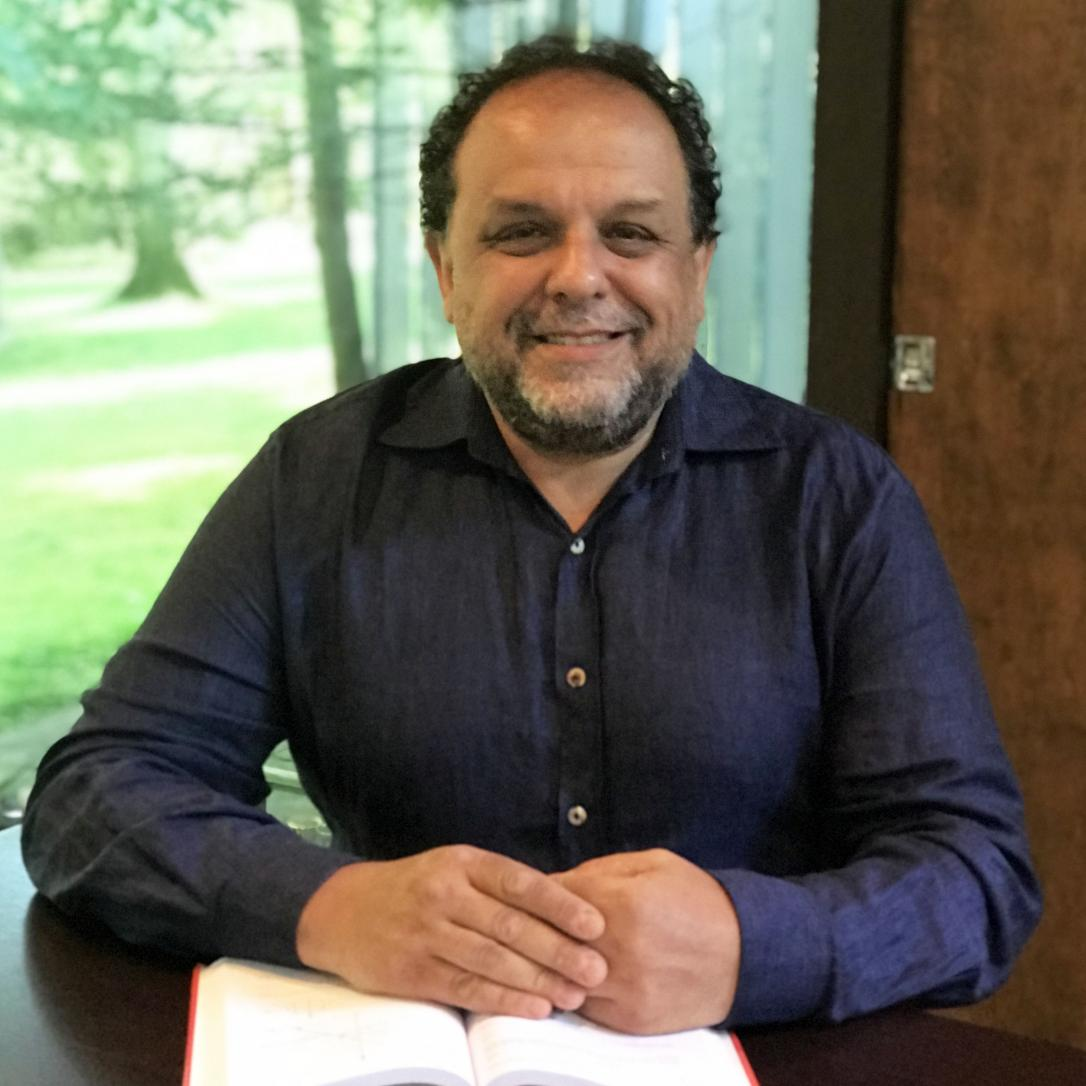}}]{Kaan Ozbay} received his B.Sc. degree from Bogazici University and his Ph.D. degree from Virginia Tech, Blacksburg. He is currently a Tenured Professor with the Department of Civil and Urban Engineering, New York University, and the Director of the C2SMART University Transportation Center. His research interests include big data analytics for smart cities, simulation of large-scale complex transportation systems, feedback-based traffic control, traffic safety, and evacuation and humanitarian inventory control. He was a recipient of the National Science Foundation (NSF) CAREER Award.
\end{IEEEbiography}

\vfill

\end{document}